\documentclass[10pt]{article}
\usepackage{epsfig,amsmath,amsthm,graphicx,amssymb,overpic}
\usepackage{epsfig,amsmath,graphicx,amssymb,overpic,cite}
\usepackage{listings,diagbox,color}
\usepackage{bm}

\usepackage{fullpage,graphicx,subfigure,mathdots,mathpazo,color}
\usepackage{amsmath,amscd,tikz,mathrsfs,cite}
\usepackage[normalem]{ulem}
\usepackage{amsmath}
\usepackage{setspace}

\def\be{\begin{equation}}
\def\ee{\end{equation}}
\def\bee{\begin{eqnarray}}
\def\ene{\end{eqnarray}}
\def\bes{\begin{subequations}}
\def\ees{\end{subequations}}

\def\v{\vspace{0.1in}}

\newcommand{\bx}{{\bm x}}

\usepackage{geometry}
\allowdisplaybreaks[4]

\begin{document}

\baselineskip=12pt
\renewcommand {\thefootnote}{\dag}
\renewcommand {\thefootnote}{\ddag}
\renewcommand {\thefootnote}{ }

\pagestyle{plain}

\begin{center}
\baselineskip=15pt \leftline{} \vspace{-.3in} {\Large \bf
Is the neural tangent kernel of PINNs deep learning general partial differential equations always convergent ?} \\[0.2in]
\end{center}

\begin{center}
Zijian Zhou and  Zhenya Yan$^{*}$\footnote{$^{*}$Corresponding author. {\it Email address}: zyyan@mmrc.iss.ac.cn}  \\[0.1in]
{\it\small Key Laboratory of Mathematics Mechanization, Academy of Mathematics and Systems Science, \\ Chinese Academy of Sciences, Beijing 100190, China \\
 School of Mathematical Sciences, University of Chinese Academy of Sciences, Beijing 100049, China} \\
\end{center}

{\baselineskip=13pt

\noindent {\bf Abstract.}\, In this paper, we study the neural tangent kernel (NTK) for general partial differential equations (PDEs) based on  physics-informed neural networks (PINNs). As we all know, the training of an artificial neural network can be converted to the evolution of NTK. We analyze the initialization of NTK and the convergence conditions of NTK during training for general PDEs. The theoretical results show that the homogeneity of differential operators plays a crucial role for the convergence of NTK.
Moreover, based on the PINNs, we validate the convergence conditions of NTK using the initial value problems of the sine-Gordon equation and the
initial-boundary value problem of the KdV equation.

\vspace{0.1in} \noindent {\it Keywords:} Deep learning; physics-informed neural networks; Neural tangent kernel; partial differential equations; convergence condition

\section{Introduction}\label{sec1}

In the past decade, artificial intelligence (AI) has witnessed widespread applications across various domains, encompassing computer vision, natural language processing, equation solving, and diverse business sectors \cite{DL1, DL2}. Some researchers began exploring the utilization of neural networks to study \textcolor{red}{PDEs} as early as the 1990s \cite{PDENN1, PDENN2} based on the approximation theory~\cite{NN0, NN1}. Recently, with the advent of remarkable advances in computational power, some scholars have renewed their focus on leveraging neural networks for PDE applications. A series of deep learning approaches have been successively proposed and achieved significant breakthroughs
in the aspect of learning PDEs. Among these, the most prevalent approach involves approximating the solutions of a PDE using neural networks, which encompasses methods, such as deep galerkin method (DGM) \cite{PINN1}, physics-informed neural networks (PINNs) \cite{PINN2}, and deep Ritz method \cite{PINN3}. Another approach is centered around approximating the solution map of a PDE using neural networks. PDE-Net \cite{map1, map2}, along with other studies \cite{map3, map4}, combine traditional numerical methods with neural networks. Similarly, DeepONet \cite{map5} and Fourier neural operator (FNO) \cite{map6}, among others \cite{map7, map8}, employ neural networks to approximate the solution map of a PDE in a black box manner.

The aforementioned neural network methods find extensive applications in diverse fields. For instance,  the transitions between two metastable states were studied in a high-dimensional probability distribution~\cite{appl1}. A fermionic neural network (FermiNet) was proposed to compute solutions to the many-electron Schr\"odinger equation~\cite{appl2}.  DeepONet was employed to predict crack paths in quasi-brittle materials~\cite{appl3}. A two-stage training method is used to deal with training the loss function that contains both equations and conservation laws \cite{chenc1}. An improved PINNs method based on Miura transformation is proposed, which realizes unsupervised learning solutions of nonlinear PDEs \cite{chenc2}. The third-order nonlinear wave equations were studied~\cite{3nls,cmkdv}. Moreover, variable coefficient PDEs are considered \cite{chenc3}. Dynamics of the one-dimensional quantum droplets are studied in \cite{chenc4}. Nonlinear dispersive equations were studied to explore peakon and periodic peakon solutions~\cite{appl4}. Additionally, Refs.~\cite{appl5,yan23,yan23-RW} delved into the study of bright solitons, breathers, and rogue wave solutions of the nonlinear Schr\"odinger-type equations.
These examples demonstrate the successful application of neural network methods across various fields.

The universality of neural networks is one of their most significant properties. The pioneering theoretical result regarding the approximation capabilities of neural networks was introduced by Cybenko in 1989 \cite{NN0}. In the context of solving \textcolor{red}{PDEs}, Physics-informed neural networks (PINNs) utilize neural networks as approximators for PDE solutions, with the first theoretical analysis presented in \cite{theo1}. Additionally, \cite{theo2} investigates over-parameterized two-layer networks and presents convergence analysis for gradient descent in the context of second-order linear PDEs.

Simultaneously, various theoretical works of neural networks were also analyzed. The NTK theory {\cite{theo3} demonstrated that the training dynamics of supervised learning models can be interpreted as kernel regression. As the width of the neural network tends to infinity, the kernel converges to a deterministic kernel. This provides a novel analytical tool for theoretical analyses of neural networks. In scenarios where the kernel approaches a constant kernel, certain pathologies can be analyzed, as explored in \textcolor{red}{\cite{theo7}}. Additionally, Ref.~\cite{theo8} considered the finite-width corrections for the limit NTK. However, NTK computation can be challenging for large-scale data. To address this, Ref.~\cite{theo9} focused on optimizing the solver for kernel methods in the context of significant scale problems. Furthermore, \cite{theo10} extended the NTK theory to convolutional neural networks.

Recently, there has been a considerable body of research that has applied the NTK  theory to diverse domains. Notably, Ref.~\cite{appl6} empirically demonstrated the consistent superiority of kernel regression employing a 14-layer CNTK over ResNet-34 trained with standard hyperparameters. This performance advantage is observed on a randomly selected subset of CIFAR-10, containing a maximum of 640 samples. Furthermore, Ref.~\cite{theo4} extended the implications of NTK theory, as presented in \cite{theo3}, to the realm of partial PINNs models. Moreover, Ref.~\cite{theo11} employed the NTK theory to establish that standard neural networks, both theoretically and practically, struggle to capture high-frequency information. Parallelly, Ref.~\cite{theo6} elucidated the characteristic of low-frequency initial learning in PINNs by leveraging the NTK framework. In this paper, we will provide a succinct overview of the primary steps involved in employing NTK to explicate the behavior of PINNs.

A fundamental difference between the NTK analysis of PINNs and the standard NTK approach lies in the incorporation of a differential operator and its associated kernel into the loss function. To be specific, Ref.~\cite{theo3} considered the following boundary condition problem in a bounded domain $\Omega\in\mathbb{R}^d$ (for the time-dependent problems, $t$ can be thought of as a part of $\bx$):
\bee\label{BP}
\begin{cases}
\mathcal{F}[q](\bx)=f(\bx),\quad \bx\in \Omega,\\
q(\bx)=g(\bx),\quad \bx\in \partial\Omega,
\end{cases}
\ene
where $\mathcal{F}$ denote a differential operator (Poisson equation and wave equation were considered in \cite{theo3}), $q(\bx)$ is the solution of $\mathcal{F}[q](\bx)=f(\bx)$ with $\bx=(x_1,x_2,...,x_d)$. For time-dependent problems, $t$ can be regarded as an additional variable of $\bx$.

The basic idea of the PINNs method \cite{PINN2} is to use a neural network to approximate the solution of the PDE. Usually, the standard choice for the fully-connected neural network (FCNN) with $L$ layers ($L-1$ hidden layers)  is defined recursively as
\begin{equation}\label{FNN}
\begin{aligned}
\bm{q}^{(0)}(\bx)=&\frac{1}{\sqrt{N_0}}\bm{W}^{(0)}\cdot\bx+\bm{b}^{(0)},\\
\bm{g}^{(i)}(\bx)=&\sigma(\bm{q}^{(i-1)}(\bx))\in\mathbb{R}^{N_i},\\
\bm{q}^{(i)}(\bx)=&\frac{1}{\sqrt{N_i}}\bm{W}^{(i)}\cdot\bm{g}^{(i)}(\bx)+\bm{b}^{(i)},\quad i=1,\cdots, L,\\
\end{aligned}
\end{equation}
where $\bm{W}^{(i)}(\in\mathbb{R}^{N_{i+1}\times N_{i}})$ and $\boldsymbol{b}^{(i)}(\in\mathbb{R}^{N_{i+1}})$ are the weight matrices to be trained, $N_i$ is the width of the $i$-th layer of neural network (i.e., number of neurons in the $i$-th layer), and $\sigma$ is a coordinate-wise activation function ($\sigma$ is usually chosen as ReLu, Sigmoid or Tanh). In the usual initialization of NTK, all the weights and biases are initialized to be independent and identically distributed (i.i.d.) as standard normal distribution $\mathcal{N}(0,1)$. The final output of neural network, $\bm{q}^{(L)}(\bx)$, can be defined as the neural network solution $\bm{q}(\bx,\bm{\theta})$, where $\bm{\theta}=\{\bm{W}^{(i)},\bm{b}^{(i)}\}_{i=0}^L$. The appropriate weight matrices of the neural network can be obtained by optimizing the PINNs loss function:
\bee\label{loss}
\mathcal{L}(\bm{\theta})=\displaystyle\frac{1}{2}\sum_{j=1}^{N_b}\left|q(\bx_b^j,\bm{\theta})-g(\bx^j_b)\right|^2 + \displaystyle\frac{1}{2}\sum_{j=1}^{N_f}\left|\mathcal{F}[q](\bx_f^j,\bm{\theta})-f(\bx^j_f)\right|^2
\ene
where $\{\bx^j_f\}_{j=1}^{N_f}$, $\{\bx^j_b, g(\bx^j_b)\}_{j=1}^{N_b}$ indicate the physical information training points and boundary data set, respectively. For the above optimization problem, if the gradient descent method (GD) with a minimal learning rate is selected, the optimization process can be converted to a gradient flow model:
\begin{equation}\label{GF}
    \frac{d\bm{\theta}}{dt}=-\nabla_{\bm{\theta}}\mathcal{L}(\bm{\theta}).
\end{equation}
After that, the gradient flow model can be rewritten as following kernel gradient descent
\bee
\begin{bmatrix}
\mathcal{F}_t[q](\bm{x}_f,\bm{\theta})  \\
q_{t}(\bm{x}_b,\bm{\theta})
\end{bmatrix}
=-\begin{bmatrix}
\mathbf{K}_{ff} & \mathbf{K}_{fb} \\
\mathbf{K}_{bf} & \mathbf{K}_{bb}
\end{bmatrix}
\begin{bmatrix}
    \mathcal{F}[q](\bm{x}_f,\bm{\theta})-f(\bm{x}_f)  \\
    q(\bm{x}_b,\bm{\theta})-g(\bm{x}_b)
\end{bmatrix},
\ene
where
\bee\label{NTKcal}
\begin{aligned}
\mathbf{K}_{ff} =& \Big(\left(\nabla_{\bm{\theta}}\mathcal{F}[q](x^i_f,\bm{\theta})\right)^T\nabla_{\bm{\theta}}
\mathcal{F}[q](x^j_f,\bm{\theta})\Big)_{i,j=1,...,N_f},\\
\mathbf{K}_{fb} = \mathbf{K}_{bf}^T =& \Big(\left(\nabla_{\bm{\theta}}\mathcal{F}[q](x^i_f,\bm{\theta})\right)^T\nabla_{\bm{\theta}}q(x_b^j,
\bm{\theta})\Big)_{i=1,...,N_f;\,j=1,...,N_b},\\
\mathbf{K}_{bb} =& \Big(\left(\nabla_{\bm{\theta}}q(x_b^i,\bm{\theta})\right)^T\nabla_{\bm{\theta}}q(x_b^j,\bm{\theta})\Big)_{i,j=1,...,N_b},
\end{aligned}
\ene
$\mathbf{K}\triangleq\begin{bmatrix}
\mathbf{K}_{ff} & \mathbf{K}_{fb} \\
\mathbf{K}_{bf} & \mathbf{K}_{bb}
\end{bmatrix}$ can be called NTK.
The main theoretical results of NTK for PINNs including:
\begin{itemize}
    \item Under the certain initialization of parameters, the initialization matrix of $\mathbf{K}(0)$ converges to a deterministic kernel $\mathbf{K^*}$ in probability \cite{theo3} when $N\rightarrow\infty$ ($N$ is the width of the hidden layer):
    \bee\label{initK}
    \mathbf{K}(0)
    \stackrel{\mathcal{P}}{\rightarrow} \mathbf{K}^*.
    \ene

    \item $\mathbf{K}$ stays asymptotically constant ($\mathbf{K}(0)$) during training (when $N\rightarrow\infty$):
    \bee    \lim\limits_{N\rightarrow\infty}\sup\limits_{t\in[0,T]}\left\|\boldsymbol{K}(t)-\boldsymbol{K}(0)\right\|_2=0.
    \ene
\end{itemize}
By consolidating the aforementioned two outcomes, we obtain the following:
\bee
\mathbf{K}(t)\approx\mathbf{K}(0)\approx\mathbf{K}^*,\quad\forall t>0.
\ene
However, it is important to note that the original theoretical findings solely focus on the Poisson equation and do not account for general PDEs. In practice, we have observed certain discrepancies in the NTK examples with respect to these results. Next, let's introduce an example:

\begin{figure}[!t]
\begin{center}
\vspace{0.05in} 
{\scalebox{0.38}[0.38]{\includegraphics{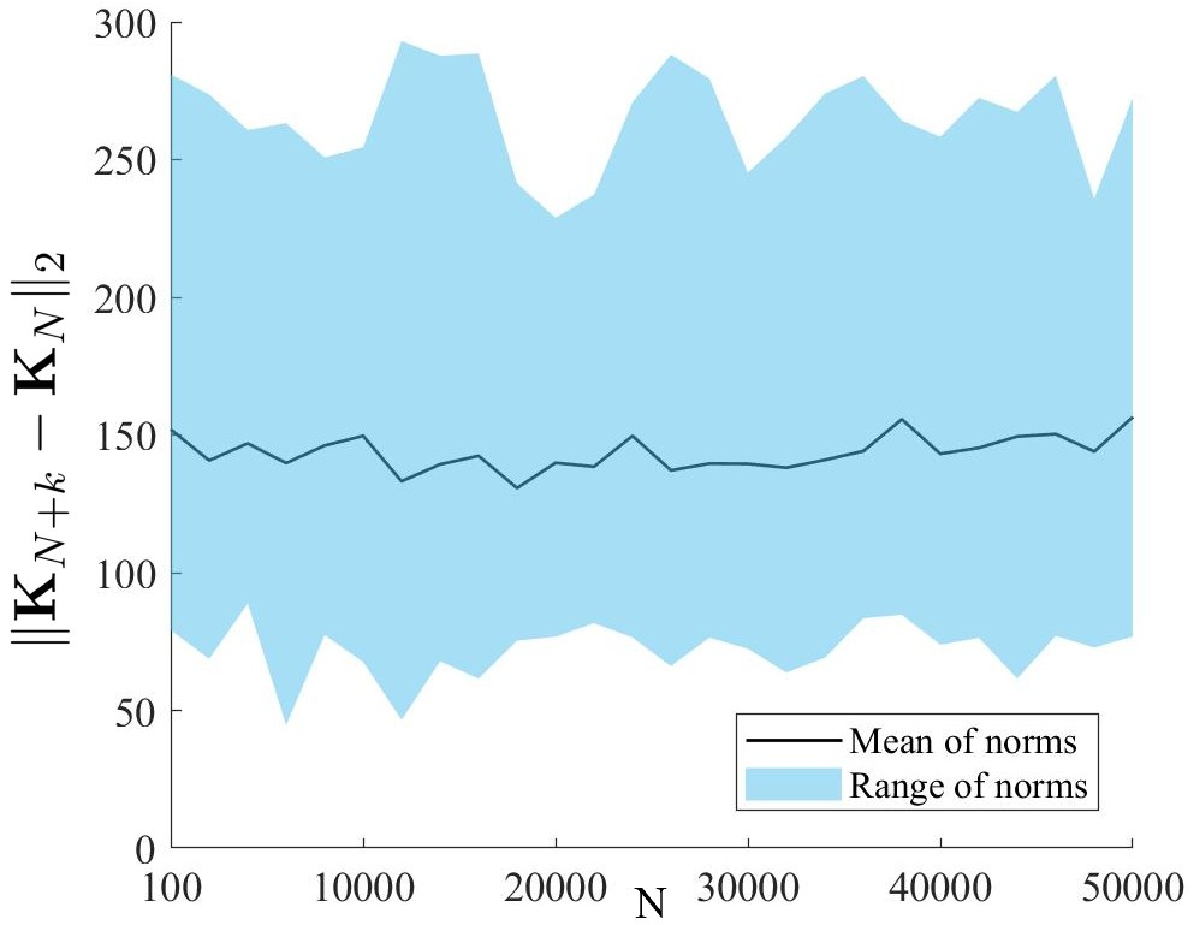}}}
\end{center}
\par
\vspace{-0.1in}
\caption{\protect\small The trends of $\|\mathbf{K}_{N+k}-\mathbf{K}_N\|_2$ of the initial value problem of the sine-Gordon equation . The blue area is the variation range of the results of 50 independent experiments, and the dark line is the mean of these experiments.}
\label{sGinit}
\end{figure}

We consider the initial value problem (IVP) of the sine-Gordon equation~\cite{soliton}:
\bee\label{IPsG}
\begin{cases}
q_{tt}(x,t)-q_{xx}(x,t)={\rm sin}(q(x,t)), \quad (x,t)\in [-5,5]\times[0,5],\\
q(x,0)=q_0(x), \quad x\in [-5,5],
\end{cases}
\ene
where $q_0(x)$ is the initial condition. An exact shock wave solution of the sine-Gordon equation would be set as the initial condition:
\bee
q(x,0)=q_0(x)=4{\rm arctan}({\rm e}^{\sqrt{2}x}).
\ene
We randomly select 50 initial sampling points and 100 spatial sampling points, denoted as $N_b=50$ and $N_f=100$, respectively. Consequently, the NTK matrix corresponds to $\mathbb{R}^{150\times150}$. To assess the convergence behavior of the initialized NTK (\ref{initK}), we investigate a series of two-layer fully connected neural networks (FCNNs) (\ref{FNN}) with a smooth activation function, specifically the hyperbolic tangent function, denoted as $\tanh(\cdot)$. The widths of these FCNNs are chosen as $\{100, 2000, 4000, 6000, ..., 50000\}$, respectively. For each width of the neural network, the weights are initialized independently and identically distributed (i.i.d.) according to a standard normal distribution, $\mathcal{N}(0,1)$, while the biases are initialized as zero. Subsequently, the initialized NTK is calculated using (\ref{NTKcal}).

Our objective is to examine whether $\mathbf{K}_N(0)$ converges to $\mathbf{K}^*$ in probability as $N$ tends to infinity. However, computing the deterministic limiting kernel $\mathbf{K}^*$ for a specific example poses challenges. Therefore, we investigate the behavior of $\|\mathbf{K}_{N+k}-\mathbf{K}_N\|_2$ instead of $\|\mathbf{K}_N-\mathbf{K}^*\|_2$, where $N$ denotes the width of the neural network and $k$ represents the interval in the sequence (in this case, $k=2000$). Although the convergence of $\|\mathbf{K}_{N+k}-\mathbf{K}_N\|_2$ and $\|\mathbf{K}_N-\mathbf{K}^*\|_2$ may not be strictly equivalent, they still provide insights into the divergence of $\|\mathbf{K}_N-\mathbf{K}^*\|_2$.

Figure \ref{sGinit} displays the trend of $\|\mathbf{K}_{N+k}-\mathbf{K}_N\|_2$. The horizontal axis $N$ corresponds to the number of neurons in the hidden layer. In order to demonstrate convergence or divergence in probability, the results are aggregated from 50 independent experiments. The blue shaded region in Fig.~\ref{sGinit} represents the outcomes of these experiments, while the black line represents the mean value across these experiments. It is evident that the sequence of initialized NTKs exhibits no discernible trend of convergence.

To gain insights into the lack of convergence of initialized NTK towards a deterministic kernel, a further discussion on NTK is warranted. The rest of this paper are organized as follows:
In Section 2, we provide a detailed analysis of NTK theory for the PDEs.
In Section 3, we present additional experimental results that shed light on the impact of the coefficient $s$ on the convergence of NTK.

\section{The neural tangent kernel for general PDEs}

In this section, we would like to consider the NTK for the boundary problems of general PDEs (GPDEs):
\bee\label{PDE}
\begin{cases}
\mathcal{F}[q](x,\bm{\theta})=f(x),\quad x\in \Omega,\\
q_{mx}(x,\bm{\theta})=g(x),\quad x\in \partial\Omega,
\end{cases}
\ene
where $\mathcal{F}[\cdot]$ is a continuous nonlinear operator,  $q_{mx}(x,\bm{\theta})$ represent some $m$-order derivative of $q(x,\bm{\theta})$ with respect with $x$. In order to consider the general case, scaling parameter $1/\sqrt{N}$ will be changed to $1/N^s$. The subsequent analysis reveals that the convergence of NTK is influenced by the convergence coefficient, denoted as $s$. It is worth noting that the two-layer FCNN is initialized in the following manner:
\bee\label{NN}
\bm{q}(x,\bm{\theta})=\frac{1}{N^s}\bm{W}^{(1)}\cdot\sigma(\bm{W}^{(0)}\cdot x+\bm{b}^{(0)})+\bm{b}^{(1)},
\ene
where $\bm{W}^{(0)}\in\mathbb{R}^{N_1\times1},\bm{b}^{(0)}\in\mathbb{R}^{N_1\times1},\bm{W}^{(1)}\in\mathbb{R}^{1\times N_1}$ and $\bm{b}^{(1)}\in\mathbb{R}$,
It is worth mentioning that the coefficient $1/N^s$ only adds to the output layer. Because the nonlinear activation function $\sigma$ is non-homogeneous, it will affect the convergence of initialized NTK (It will be discussed in detail in the proof of Theorem 2.2). The loss function is chosen as
\bee \label{loss1}
\mathcal{L}(\bm{\theta})=\alpha\mathcal{L}_f(\bm{\theta})+\beta\mathcal{L}_b(\bm{\theta})
=\displaystyle\frac{\alpha}{2}\sum_{i=1}^{N_f}\left(\mathcal{F}[q](x^i_f,\bm{\theta})-f(x^i_f)\right)^2+\displaystyle\frac{\beta}{2}\sum_{i=1}^{N_b}\left(q_{mx}(x_b^i,\bm{\theta})-g(x_b^i)\right)^2.
\ene
where $\alpha,\, \beta$ are added to balance the loss of different terms.

In the same way in Sec.~\ref{sec1}, the optimization process can be changed to the kernel gradient descent.

\noindent {\bf Lemma 2.1.} Given the data points $\{x^i_f,f(x^i_f)\}^{N_f}_{i=1}$, $\{x^i_b,g(x^i_b)\}^{N_b}_{i=1}$ and the gradient flow (\ref{GF}). $\mathcal{F}[q](x,\bm{\theta})\in\mathbb{R}^{N_f\times1}$, obey the following matrix evolution equation
\bee\label{kernel}
\begin{bmatrix}
\mathcal{F}_t[q](x_f,\bm{\theta})  \\
q_{mx,t}(x_b,\bm{\theta})
\end{bmatrix}
=-\begin{bmatrix}
\alpha\mathbf{K}_{ff} & \beta\mathbf{K}_{fb} \\
\alpha\mathbf{K}_{bf} & \beta\mathbf{K}_{bb}
\end{bmatrix}
\begin{bmatrix}
    \mathcal{F}[q](x_f,\bm{\theta})-f(x_f)  \\
    q_{mx}(x_b,\bm{\theta})-g(x_b)
\end{bmatrix}
\ene

where
\bee
\begin{aligned}
\mathbf{K}_{ff} =& \Big(\left(\nabla_{\bm{\theta}}\mathcal{F}[q](x^i_f,\bm{\theta})\right)^T\nabla_{\bm{\theta}}\mathcal{F}[q](x^j_f,
\bm{\theta})\Big)_{i,j=1,...,N_f}, \\
\mathbf{K}_{fb} = \mathbf{K}_{bf}^T =& \Big(\left(\nabla_{\bm{\theta}}\mathcal{F}[q](x^i_f,
\bm{\theta})\right)^T\nabla_{\bm{\theta}}q_{mx}(x_b^j,\bm{\theta})\Big)_{i=1,...,N_f;\, j=1,...,N_b},\\
\mathbf{K}_{bb} =& \Big(\left(\nabla_{\bm{\theta}}q_{mx}(x_b^i,\bm{\theta})\right)^T\nabla_{\bm{\theta}}q_{mx}(x_b^j,\bm{\theta})\Big)_{i,j=1,...,N_b}.
\end{aligned}
\ene

\v\noindent {\it Proof.} The proof of lemma 2.1 is given in Appendix A. $\qedsymbol$ \v

When employing gradient descent (GD) with a learning rate that tends to 0, Lemma 2.1 establishes that the training process can be reformulated as a kernel gradient descent problem. Consequently, the investigation of training processes can be conducted by analyzing the behavior of the kernel matrix.

The first theoretical finding affirms that the initialized NTK converges, in probability, to a deterministic kernel matrix as the width of the neural network, denoted as $N$, approaches infinity.

\v\noindent  {\bf Theorem 2.2.} For the kernel (\ref{kernel}) of the boundary problem of PDE, where $\mathcal{F}$ is a continuous differential operator, when the width of the neural network, denoted as $N$, tends to infinity, the kernel associated with the neural network (\ref{NN}) converges in probability to the following deterministic limiting kernel:

\bee
\mathbf{K}(0)=\begin{bmatrix}
\alpha\mathbf{K}_{ff}(0) & \beta\mathbf{K}_{fb}(0) \\
\alpha\mathbf{K}_{bf}(0) & \beta\mathbf{K}_{bb}(0)
\end{bmatrix}
\stackrel{\mathcal{P}}{\rightarrow} \mathbf{K}^*,
\ene
where $\mathbf{K}_{ff}(0),\mathbf{K}_{fb}(0),\mathbf{K}_{bf}(0),\mathbf{K}_{bb}(0)$ are defined in Lemma 2.1, and $\stackrel{\mathcal{P}}{\rightarrow}$ represent convergence by probability.

\v \noindent {\it Proof.} The proof of Theorem 2.2 is given in Appendix B. \v

\noindent {\bf Remark 1.} The crucial aspect determining kernel convergence is the balance between the number of neurons and the convergence coefficient, represented by $N^{-s}$. The convergence of $\mathbf{K}_{bb}$ occurs in probability when $s\geq\frac{1}{2}$. In the case where $F_0[q,q_x,q_{xx},...,q_{nx}]\cdot F_0[\hat{q},\hat{q}_x,\hat{q}_{xx},...,\hat{q}_{nx}]\neq0$ ($F_0[q,q_x,q_{xx},...,q_{nx}]\neq0$), or if there exists a non-homogeneous monomial in the polynomial $\{F_i[q,q_x,q_{xx},...q_{nx}]\}_{i=1}^n$ (referred to as case A), the convergence of $\mathbf{K}_{ff}$ ($\mathbf{K}_{bf}$) occurs in probability when $s\geq1$. On the other hand, if $F_0[q,q_x,q_{xx},...,q_{nx}]\cdot F_0[\hat{q},\hat{q}_x,\hat{q}_{xx},...,\hat{q}_{nx}]=0$ ($F_0[q,q_x,q_{xx},...,q_{nx}]=0$), or if every monomial in the polynomials $\{F_i[q,q_x,q_{xx},...,q_{nx}]\}_{i=1}^n$ is homogeneous (referred to as case B), the convergence of $\mathbf{K}_{ff}$ ($\mathbf{K}_{bf}$) occurs in probability when $s\geq s_1$ (as defined in Eq.~(\ref{s1})) ($s\geq s_2$ as defined in
Eq.~(\ref{s2})).

In other words, in case A, $\mathbf{K}$ convergence by probability when $s\geq1$. In case B, $\mathbf{K}$ convergence by probability when $s\geq s_1(>s_2)$. This means the past initialization coefficient $N^{-1/2}$ does not universally guarantee convergence. It depends on the homogeneity of $\{F_i[q,q_x,q_{xx},...q_{nx}](x,\bm{\theta})\}_{i=1}^n$ and the value of $F_0[q,q_x,q_{xx},...q_{nx}](x,\bm{\theta})$. In fact, in the examples in the first and third sections, we observe instances where the initialized NTK diverges.

\v\noindent {\bf Remark 2.} The significance of the non-homogeneous term $F_i$ in the analysis of Theorem 2 is evident. In a multi-layer network, the nonlinear activation function also exhibits non-homogeneity. When the convergence coefficient $N^{-s}$ is introduced to the hidden layer, the balance between the number of neurons and the convergence coefficient becomes challenging.

\v Motivated by Theorem 2.2, a comprehensive examination of the NTK range during training is undertaken. Remarkably, the NTK remains a constant matrix (referred to as the initialized NTK) throughout the training process when $N\rightarrow\infty$ and $s>1/4$.

\v\noindent {\bf Theorem 2.3} For the loss function (\ref{loss}), if the following assumptions are satisfied for any $T>0$:

(i) For $t\in T$, all parameters of the network are uniformly bounded, i.e., there exists a constant $C>0$ (independent on $N$) such that
\begin{equation}
\sup _{t \in[0, T]}\|\boldsymbol{\bm{\theta}}(t)\|_{\infty} \leq C,
\end{equation}

(ii) The derivatives of the equation are uniformed bounded, i.e. there exists a constant $C>0$ (independent on $n$) such that
\begin{equation}
\sup _{i \in\{0,1,...,n\}}\left\|F_i[q,q_x,q_{xx},...,q_{nx}]\right\|_{\infty} \leq C,
\end{equation}

(iii) There exists a constant $C>0$,
\begin{equation}
\begin{aligned}
&\int_{0}^{T}\left|\sum_{i=1}^{N_{b}}\left(\frac{\partial^m q}{\partial x^m}\left(x_{b}^{i}, \boldsymbol{\bm{\theta}}(\tau)\right)-f\left(x_{b}^{i}\right)\right)\right| d \tau \leq C, \\
&\int_{0}^{T}\left|\sum_{i=1}^{N_{f}}\left(\mathcal{F}[q](x^j_f,\bm{\theta}(\tau))-g(x^i_b)\right)\right| d \tau \leq C,
\end{aligned}
\end{equation}

(iv) The activation function $\sigma$ is smooth and its $k$-order derivatives are bounded (i.e. $|\sigma^{(i)}|\leq C,\, 0\leq i\leq k$),

then when $s>1/4$ we have
\bee
\lim\limits_{N\rightarrow\infty}\sup\limits_{t\in[0,T]}\left\|\boldsymbol{K}(t)-\boldsymbol{K}(0)\right\|_2=0.
\ene

\v\noindent {\it Proof.} The proof of Theorem 2.3 is given in Appendix C. \v



\v\noindent {\bf Remark 3.} Condition (ii) imposes the requirement that the derivatives of the equation are uniformly bounded. This condition shows that if $F_i[q,q_x,q_{xx},...,q_{nx}]$ is unbounded, the convergence of $\boldsymbol{K}(t)$ may not be guaranteed. One example of such an unbounded term is the logarithmic function.

\v\noindent {\bf Remark 4.} Theoretical analysis indicates that the Neural Tangent Kernel (NTK) remains a constant matrix when $s>1/4$. This finding extends the range of initialized neural networks, which was previously established for the case of $s=1/2$. The convergence of the NTK during training for values of $s<1/2$ is verified in next section. In next section, we present two examples where the NTK remains a constant matrix throughout the training process for $s=1/4$.

In the preceding analysis, we provide a detailed discussion of the original conclusion, introducing relaxations or additional constraints to the theorem. These derived conclusions serve to expand or restrict the applicability of the NTK theory. The subsequent numerical examples serve to substantiate the validity of such an analysis.

\section{Some examples}

In this section, we present two illustrative examples to elucidate the influence of the coefficient $s$ on the convergence of the initialized NTK and the NTK during training. Specifically, we examine the initial value problem of the non-homogeneous nonlinear sine-Gordon equation represented in Sec. \ref{sG}, and the initial boundary value problem of the homogeneous nonlinear KdV equation depicted in Sec. \ref{KdV}.

\begin{figure}[!t]
\begin{center}
\vspace{0.05in} 
{\scalebox{0.6}[0.6]{\includegraphics{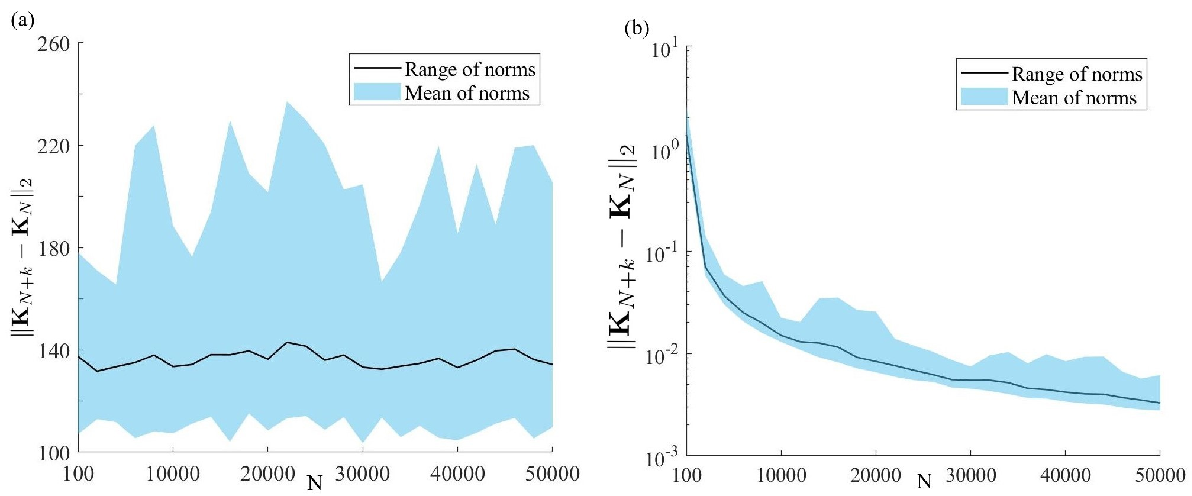}}}
\end{center}
\par
\vspace{-0.05in}
\caption{\protect\small The trends of $\|\mathbf{K}_{N+k}-\mathbf{K}_N\|_2$ of the initial value problem of the sine-Gordon equation (\ref{IPsG}) when (a) $s=0.5$, (b) $s=1$. The blue area is the variation range of the results of 50 independent experiments, and the black line denotes the mean value of these experiments.}
\label{sGinit1}
\end{figure}

\subsection{NTK convergence of sine-Gordon equation}\label{sG}

In the first experiment, we continue to examine the initial value problem given by Eq.~(\ref{IPsG}). As discussed in Remark 2 of Theorem 2.2, the convergence of the NTK is influenced by the nonlinear activation function when the convergence coefficient $s$ is not sufficiently large. Consequently, in the subsequent experiments, we only apply the convergence coefficient to the output layer (\ref{NN}). We employ the same series of neural networks as in Section 2 to validate the theoretical findings. Given that the initialized NTK converges probabilistically to a deterministic kernel, we independently run the program 50 times and subsequently plot the average and range of the results in Fig.\ref{sGinit1}.

Figure \ref{sGinit1} presents the experimental results for two different values of $s$: $s=0.5$ (Fig.~\ref{sGinit1}a) and $s=1$ (Fig.~\ref{sGinit1}b). It is evident that the initialized NTK diverges when $s=0.5$, but converges when $s=1$. This observation highlights the influence of the parameter $s$ on the convergence and divergence of the initialized NTK. Specifically, it demonstrates that only when $s$ is sufficiently large, the initialized NTK for the sine-Gordon equation is guaranteed to converge to a deterministic kernel. As analyzed in Theorem 2.2, the presence of non-homogeneous terms such as sin($\cdot$) leads to the divergence of the initialized NTK.

We further investigate the convergence of NTK during training with a convergence coefficient of $s=1/4$. A series of two-layer FCNNs with varying widths, ranging from ${200, 400, \ldots, 5000}$, are trained using the standard gradient descent method with the fixed weighted Loss function (\ref{loss1}). For each width of the FCNN, we train the models for different numbers of steps, specifically 5000, 10000, 20000, and 50000 steps. The results are plotted as separate curves in Figure \ref{sGtrain}. To calculate the loss function, we randomly select 50 initial sampling points and 100 spatial sampling points. Each point on the graph in Fig.~\ref{sGtrain} represents the upper bound of the 2-norm difference between the NTK and the initial NTK over the entire training process.

Figure \ref{sGtrain} presents the training results under these conditions. As the width $N$ increases, the quantity $\sup\limits_{t\in[0,n]}\|\mathbf{K}_N(t)-\mathbf{K}_N(0)\|_2$ consistently decreases, regardless of the maximum number of training steps. This observation indicates that the NTK remains a constant matrix throughout the training process when $s=1/4$. Consequently, this finding suggests that NTK theory can be applied to a broader range of FCNN initialization scenarios.

\begin{figure}[!t]
\begin{center}
{\scalebox{0.3}[0.3]{\includegraphics{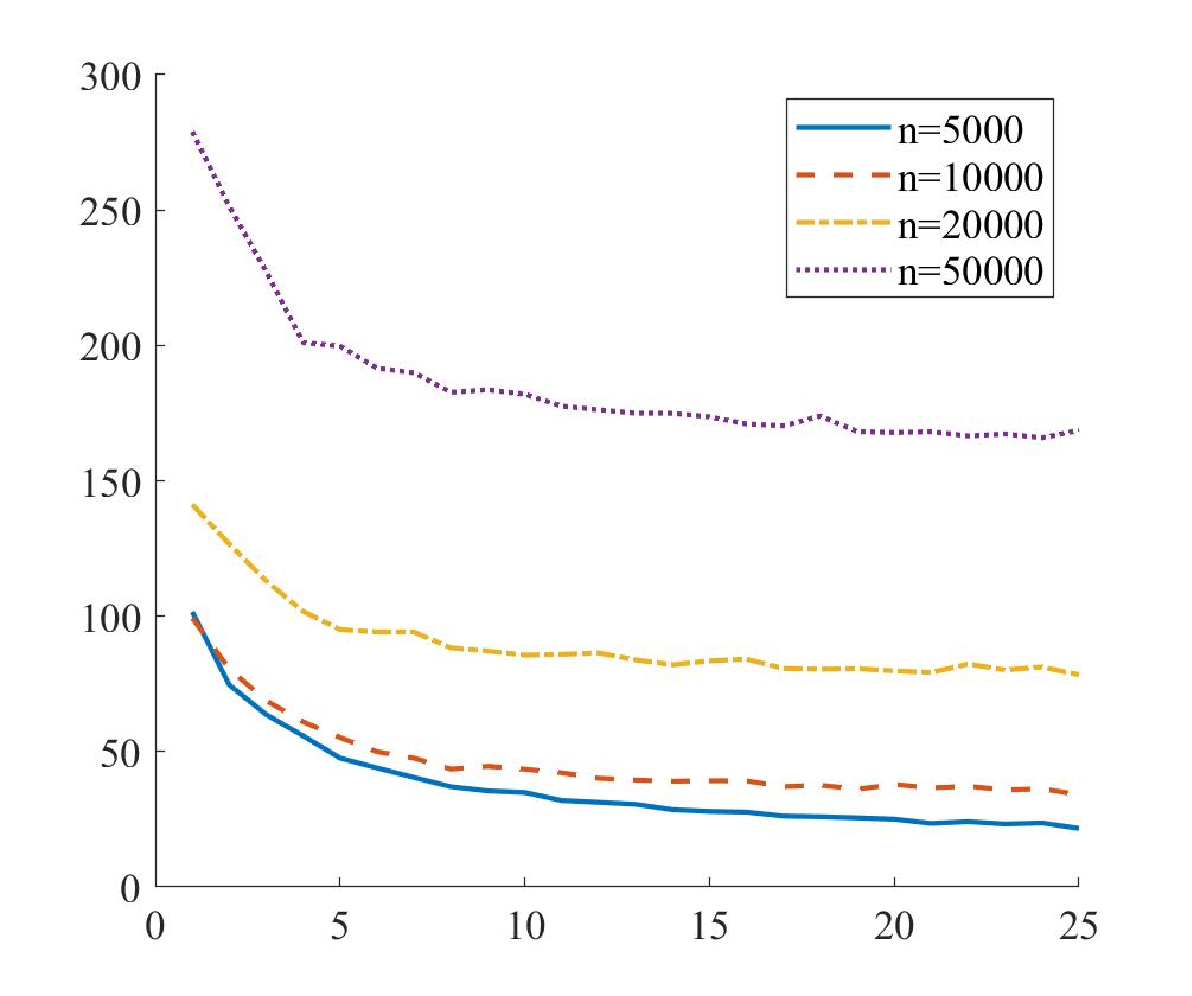}}}
\end{center}
\vspace{-0.2in}
\caption{\protect\small The trends of $\sup\limits_{t\in[0,n]}\|\mathbf{K}_N(t)-\mathbf{K}_N(0)\|_2$ of the initial value problem of the sine-Gordon equation (\ref{IPsG}) when $s=1/4$, where $n$ represents the maximal training step and $N$ the width of the neural networks.}
\label{sGtrain}
\end{figure}

\subsection{NTK convergence of KdV equation}\label{KdV}

In the second experiment, we investigate the initial boundary value problem of the homogeneous nonlinear KdV equation~\cite{soliton}:
\bee\label{IBPKdV}
\begin{cases}
q_{t}(x,t)+6q(x,t)q_{x}(x,t)+q_{xxx}(x,t)=0,  \quad (x,t)\in (-5, 5)\times (0, 5),\\
q(-5,t)=q(5,t),  \quad t\in [0, 5],\\
q(x,0)=q_0(x),  \quad x\in [-5, 5].
\end{cases}
\ene
Periodic boundary conditions are imposed by incorporating a boundary loss term into the overall Loss function (\ref{loss1}):
\bee
\begin{aligned}
\mathcal{L}(\bm{\theta})=&\displaystyle\frac{1}{2}\sum_{j=1}^{N_i}\left|q(x_i^j,0,\bm{\theta})-q_0(\bx^j_i)\right|^2 + \displaystyle\frac{1}{2}\sum_{j=1}^{N_b}\left|q(-5,t_b^j,\bm{\theta})-q(5,t_b^j,\bm{\theta})\right|^2 \\
&+ \displaystyle\frac{1}{2}\sum_{j=1}^{N_f}\left|\mathcal{F}[q](x_f^j,t^j_f,\bm{\theta})-f(x^j_f,t^j_f)\right|^2.
\end{aligned}
\ene
And the diagonal of the kernel (\ref{kernel}) will be made up of $\mathbf{K}_{ii}$, $\mathbf{K}_{bb}$ and $\mathbf{K}_{ff}$. We take an exact soliton solution as the initial value of this problem:
\bee
q_0(x)=2b^2{\rm sech}(bx)^2, \quad x\in [-5,5].
\ene
and 50 initial sampling points, 50 boundary sampling points and 100 spatial sampling points are ramdomly selected to computing the Loss function (i.e. $\mathbf{K}\in \mathbb{R}^{200\times200}$).

Fig.\ref{KdVinit} displays the convergence of initialized NTK of the KdV equation. A series of two-layer FCNNs with a width of $\{100, 2000, 4000, ..., 50000\}$ are used to verify the conclusions of section 3. As N increases, NTK tends to converge at $s=1$ and tends to diverge at $s=0.5$. Since there is a product term $6uu_x$ in the KdV equation, $F_i$ (\ref{maxord}) is not all equal to constant. And by the analysis in the proof of Theorem 2.2, $s_1=3/4$ (\ref{s1}) and $s_2=2/3$ (\ref{s2}). Only when $s\geq {\rm max}\{s_1,s_2\}=3/4$, initialized NTK is convergent.

Figure \ref{KdVinit} depicts the convergence behavior of the initialized NTK for the KdV equation. A series of two-layer FCNNs with varying widths, namely $\{100, 2000, 4000, ..., 50000\}$, are employed to investigate the conclusions outlined in Section 2. The trend observed is that as the width, N, increases, the NTK tends to converge when $s=1$, while it tends to diverge when $s=0.5$. This behavior can be attributed to the presence of the product term $6uu_x$ in the KdV equation, which results in non-constant values for $F_i$ given by Eq.~(\ref{maxord}). Based on the analysis presented in the proof of Theorem 2.2, it can be determined that $s_1=3/4$ given by Eq.~(\ref{s1}) and $s_2=2/3$ given by Eq.~(\ref{s2}). Consequently, only when $s\geq {\rm max}{s_1,s_2}=3/4$, the initialized NTK exhibits convergence.

\begin{figure}[!t]
\begin{center}
{\scalebox{0.6}[0.6]{\includegraphics{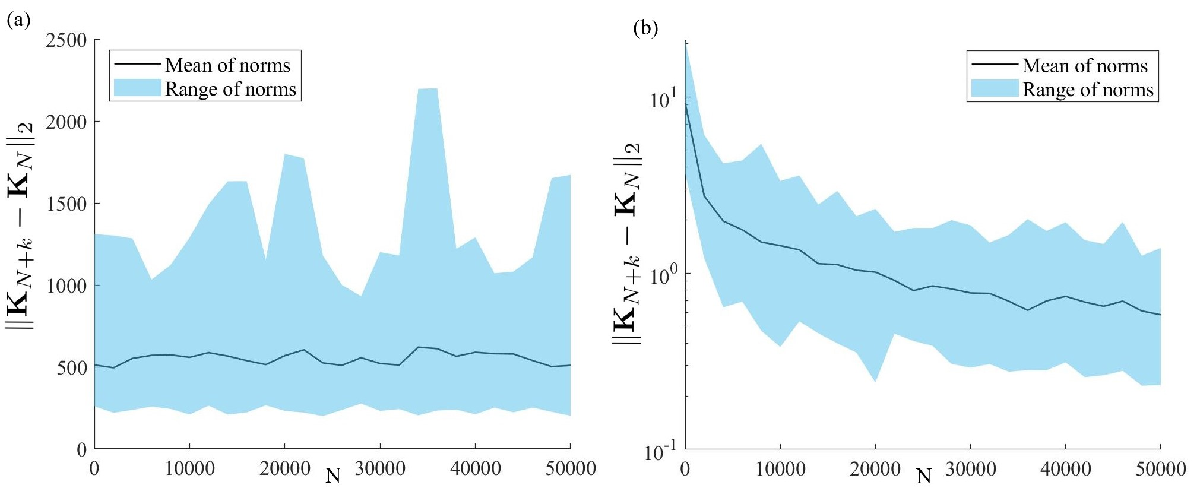}}}
\end{center}
\par
\vspace{-0.05in}
\caption{\protect\small The trends of $\|\mathbf{K}_{N+k}-\mathbf{K}_N\|_2$ of the initial value problem of the KdV equation (\ref{IBPKdV}) when (a) $s=0.5$, (b) $s=1$. The blue area is the variation range of the results of 50 independent experiments, and the black line is the mean value of these experiments.}
\label{KdVinit}
\end{figure}

Figure \ref{KdVtrain} illustrates the convergence of NTK during the training process. Each FCNN width is trained using the standard gradient descent method with a fixed step size of $10^{-5}$. The difference between the NTK and initial NTK tends to zero at different training steps. In this experiment, the convergence coefficient $s$ is set to $0.3$, thereby validating the result presented in Theorem 3.

\begin{figure}
\begin{center}
\vspace{0.05in} 
{\scalebox{0.31}[0.31]{\includegraphics{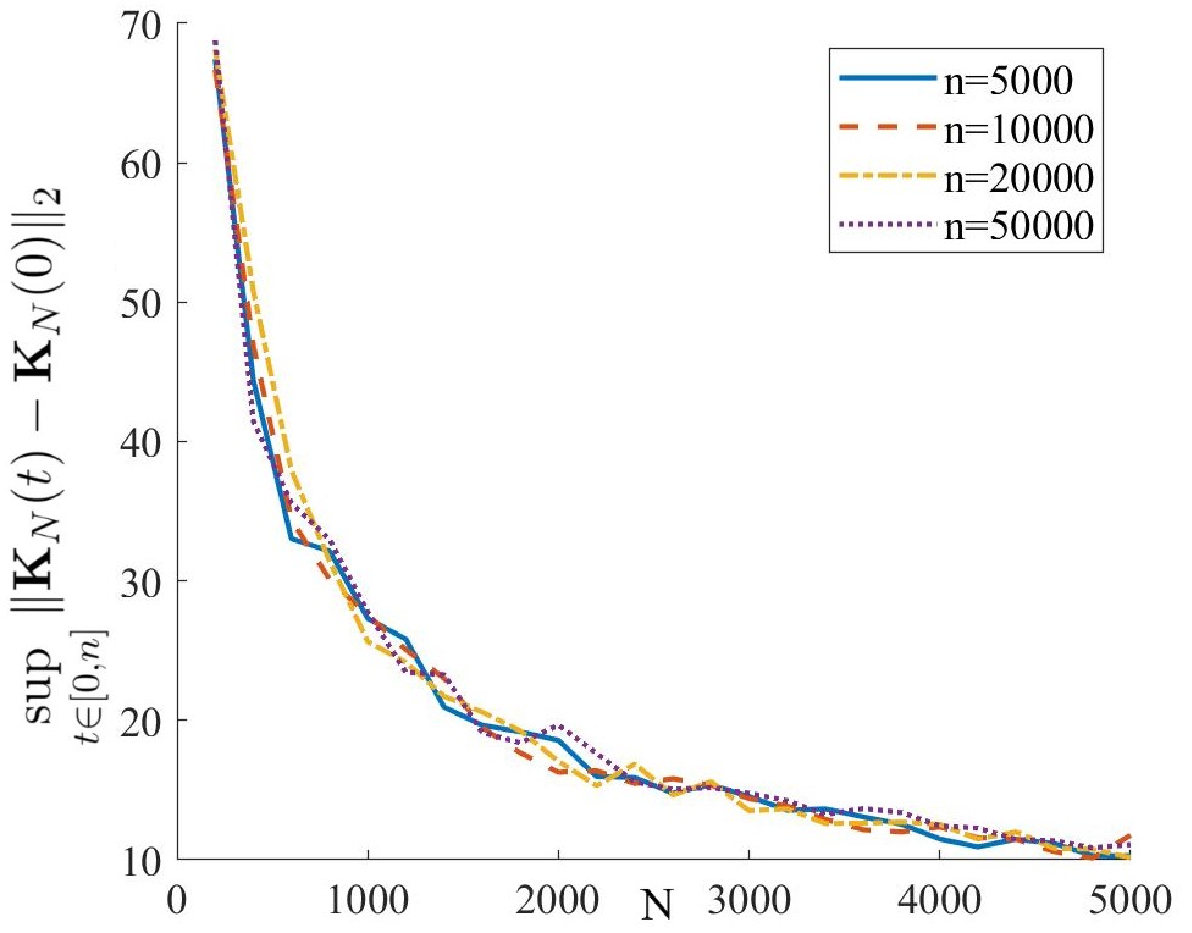}}}
\end{center}
\par
\vspace{-0.1in}
\caption{\protect\small The trends of $\sup\limits_{t\in[0,n]}\|\mathbf{K}_N(t)-\mathbf{K}_N(0)\|_2$ of the initial value problem of the KdV equation when $s=0.2$, where $n$ represents the maximal train step and $N$  the width of the neural networks.}
\label{KdVtrain}
\end{figure}

\section{Conclusions and discussions}

In conclusion, we have found some divergent cases of initialized NTK under normal conditions. The convergence of NTK is discussed in more detail, and some results were changed. We add some restrictions in the convergence of initialized NTK (Theorem 2.2), and the convergence condition of NTK during training is relaxed (Theorem 2.3). These findings contribute to a clearer understanding of NTK theory for the general PINNs models, enabling its application to a broader range of problem domains.



\v\v \noindent {\bf Appendix A. \, Proof of Lemma 2.1}

\v\noindent {\it Proof.} In case A, the loss function is
\bee
\mathcal{L}(\bm{\theta})=\mathcal{L}_f(\bm{\theta})+\mathcal{L}_b(\bm{\theta})
=\displaystyle\frac{\alpha}{2}\sum_{i=1}^{N_f}\left(\mathcal{F}[q](x^i_f,\bm{\theta})-f(x^i_f)\right)^2
+\displaystyle\frac{\beta}{2}\sum_{i=1}^{N_b}\left(q_{mx}(x_b^i,\bm{\theta})-g(x_b^i)\right)^2,
\ene
where $\alpha,\, \beta$ are weights.

And the gradient flow is
\begin{equation}
    \frac{d\bm{\theta}}{dt} = -\nabla_{\bm{\theta}}\mathcal{L}(\bm{\theta}) =
    -\alpha\sum_{i=1}^{N_f}\left(\mathcal{F}[q](x^i_f,\bm{\theta})-f(x^i_f)\right)\nabla_{\bm{\theta}}\mathcal{F}[q](x^i_f,\bm{\theta})
    -\beta\sum_{i=1}^{N_b}\left(q_{mx}(x_b^i,\bm{\theta})-g(x_b^i)\right)\nabla_{\bm{\theta}}q_{mx}(x_b^i,\bm{\theta}).
\end{equation}
Which is a $(3N+1)$-dimensional equation set. For $i=1,...,N_f$
\bee
\begin{aligned}
    \frac{d\mathcal{F}[q](x^i_f,\bm{\theta})}{dt}
    =& \nabla_{\bm{\theta}}\mathcal{F}[q](x^i_f,\bm{\theta})^T\cdot \frac{d\bm{\theta}}{dt} \\
    =& -\nabla_{\bm{\theta}}\mathcal{F}[q](x^i_f,\bm{\theta})^T\cdot \bigg[\sum_{j=1}^{N_f}\alpha\left(\mathcal{F}[q](x^j_f,\bm{\theta})-f(x^j_f)\right)
    \nabla_{\bm{\theta}}\mathcal{F}[q](x^j_f,\bm{\theta})\\
    & +\sum_{j=1}^{N_b}\beta\left(q_{mx}(x_b^j,\bm{\theta})-g(x^j_b)\right)\nabla_{\bm{\theta}}q_{mx}(x_b^j,\bm{\theta})\bigg] \\
    =&-\sum_{j=1}^{N_f}\alpha\left(\mathcal{F}[q](x^j_f,\bm{\theta})-f(x^j_f)\right)\left(\nabla_{\bm{\theta}}\mathcal{F}[q](x^i_f,\bm{\theta})\right)^T\nabla_{\bm{\theta}}\mathcal{F}[q](x^j_f,\bm{\theta})\\
    &-\sum_{j=1}^{N_b}\beta\left(q_{mx}(x_b^j,\bm{\theta})-g(x^j_b)\right)\left(\nabla_{\bm{\theta}}\mathcal{F}[q](x^i_f,\bm{\theta})
    \right)^T\nabla_{\bm{\theta}}q_{mx}(x_b^j,\bm{\theta}), \\
\end{aligned}
\ene
that is,
\bee
\frac{d\mathcal{F}[q](\bm{x},\bm{\theta})}{dt} =
-\begin{bmatrix}
    \mathbf{K}_{ff}, & \mathbf{K}_{fb}
\end{bmatrix}
\begin{bmatrix}
    \mathcal{F}[q](\bm{x}_f,\bm{\theta})-f(\bm{x}_f)  \\
    q_{mx}(\bm{x}_b,\bm{\theta})-g(\bm{x}_b)
\end{bmatrix}.
\ene
Similarly, we have
\bee
\frac{dq_{mx}(\bm{x}_b,\bm{\theta})}{dt} =
-\begin{bmatrix}
    \mathbf{K}_{bf}, & \mathbf{K}_{bb}
\end{bmatrix}
\begin{bmatrix}
    \mathcal{F}[q](\bm{x}_f,\bm{\theta})-f(\bm{x}_f)  \\
    q_{mx}(\bm{x}_b,\bm{\theta})-g(\bm{x}_b)
\end{bmatrix}.
\ene
$\qedsymbol$

\v \noindent {\bf Appendix B. \, Proof of Theorem 2.2}\label{Th1}

\v\noindent {\it Proof.} We divide the proof into 3 parts: convergence of $\mathbf{K}_{ff}$, $\mathbf{K}_{bb}$ and $\mathbf{K}_{bf}$.
\\
(1) We first consider the convergence of  $\mathbf{K}_{ff}$: We assume $\mathcal{F}[q](x,\bm{\theta})=F[q,q_x,q_{xx},...q_{nx}](x,\bm{\theta})=f(x)$, where $\mathcal{F}[q]$ is a continuous linear or nonlinear function and $n$ is the number of the max order derivative. The partial derivative of $F[q,q_x,q_{xx},...q_{nx}](x,\bm{\theta})$ with respect to $\bm{\theta}$ can be written as
\bee
\begin{aligned}
\frac{\partial F[q,q_x,q_{xx},...q_{nx}](x,\bm{\theta})}{\partial\bm{\theta}}
=&\sum\limits_{i=0}^nF_i[q,q_x,q_{xx},...q_{nx}]\frac{\partial q_{ix}(x,\bm{\theta})}{\partial\bm{\theta}},\\
\end{aligned}
\ene
where $F_i[q,q_x,q_{xx},...q_{nx}](x,\bm{\theta})$ represent the derivative with respect to $q_{ix}$. Then for any two data points $x,\hat{x}$, we have
\bee
\begin{aligned}
\mathbf{K}_{ff}=&\Big\langle\nabla_{\bm{\theta}}\mathcal{F}[q](x,\bm{\theta}), \nabla_{\bm{\theta}}\mathcal{F}[q](\hat{x},\bm{\theta})\Big\rangle\\
=&\bigg(\sum\limits_{i=0}^nF_i[q,q_x,q_{xx},...,q_{nx}]\frac{\partial q_{ix}(x,\bm{\theta})}{\partial\bm{\theta}}\bigg)^T\bigg(\sum\limits_{j=0}^nF_j[\hat{q},\hat{q}_x,\hat{q}_{xx},...,\hat{q}_{nx}]\frac{\partial q_{jx}(\hat{x},\bm{\theta})}{\partial\bm{\theta}}\bigg)\\
=&\sum\limits_{k=0}^{3N+1}\bigg(\sum\limits_{i=0}^nF_i[q,q_x,q_{xx},...,q_{nx}]\frac{\partial q_{ix}(x,\bm{\theta})}{\partial\bm{\theta}_k}\bigg)^T\bigg(\sum\limits_{j=0}^nF_j[\hat{q},\hat{q}_x,\hat{q}_{xx},...,\hat{q}_{nx}]\frac{\partial q_{jx}(\hat{x},\bm{\theta})}{\partial\bm{\theta}_k}\bigg)\\
=&\sum\limits_{k=0}^{3N}\bigg(\sum\limits_{i=0}^nF_i[q,q_x,q_{xx},...,q_{nx}]\frac{\partial q_{ix}(x,\bm{\theta})}{\partial\bm{\theta}_k}\bigg)^T\bigg(\sum\limits_{j=0}^nF_j[\hat{q},\hat{q}_x,\hat{q}_{xx},...,\hat{q}_{nx}]\frac{\partial q_{jx}(\hat{x},\bm{\theta})}{\partial\bm{\theta}_k}\bigg)\\
&+\bigg(\sum\limits_{i=0}^nF_i[q,q_x,q_{xx},...,q_{nx}]\frac{\partial q_{ix}(x,\bm{\theta})}{\partial\bm{b}^{(1)}}\bigg)^T\bigg(\sum\limits_{j=0}^nF_j[\hat{q},\hat{q}_x,\hat{q}_{xx},...,\hat{q}_{nx}]\frac{\partial q_{jx}(\hat{x},\bm{\theta})}{\partial\bm{b}^{(1)}}\bigg)\\
=&\sum\limits_{k=0}^{3N}\bigg(\sum\limits_{i=0}^nF_i[q,q_x,q_{xx},...,q_{nx}]\frac{\partial q_{ix}(x,\bm{\theta})}{\partial\bm{\theta}_k}\bigg)^T\bigg(\sum\limits_{j=0}^nF_j[\hat{q},\hat{q}_x,\hat{q}_{xx},...,\hat{q}_{nx}]\frac{\partial q_{jx}(\hat{x},\bm{\theta})}{\partial\bm{\theta}_k}\bigg)\\
&+F_0[q,q_x,q_{xx},...,q_{nx}]F_0[\hat{q},\hat{q}_x,\hat{q}_{xx},...,\hat{q}_{nx}],
\end{aligned}
\ene
where $\bm{\theta}=\{\bm{W}^{(0),T},\bm{b}^{(0),T},\bm{W}^{(1)},\bm{b}^{(1)}\}\in\mathbb{R}^{3N+1}$. Recall that
\bee
\begin{aligned}
q(x, \boldsymbol{\bm{\theta}})=&\frac{1}{N^s} \sum_{k=1}^{N}\boldsymbol{W}^{(1)}_k\sigma_k+\boldsymbol{b}^{(1)},\\
q_{ix}(x, \boldsymbol{\bm{\theta}})=&\frac{1}{N^s} \sum_{k=1}^{N}\boldsymbol{W}^{(1)}_k\sigma^{(i)}_k\boldsymbol{W}^{(0)i}_k,\quad i=1,...,n,
\end{aligned}
\ene
where $\sigma^{(i)}_k$ represent $\sigma^{(i)}\left(\boldsymbol{W}^{(0)}_kx+\boldsymbol{b}^{(0)}_k\right)$. In order to discuss the convergence of $\mathbf{K}_{ff}$, the coefficient $\frac{1}{N^s}$ will be discussed.

(a) As $F_0[q,q_x,q_{xx},...,q_{nx}]F_0[\hat{q},\hat{q}_x,\hat{q}_{xx},...,\hat{q}_{nx}]\neq0$ or exist a monomial in  $\{F_i[q,q_x,q_{xx},...q_{nx}](x,\bm{\theta})\}_{i=1}^n$ is non-homogeneous.

{\bf Definition}:\, (homogeneous) If $F_i[q,q_x,q_{xx},...q_{nx}]$ is a homogeneous operator,  there exist $\{t_{i}\}_{i=0}^{n}\subset\mathbb{N}$, such that for any $\{m_i\}_{i=0}^{n}\subset\mathbb{R}$:
\bee\label{homo}
F[m_0q,m_1q_x,m_2q_{xx},...,m_nq_{nx}]=m_0^{t_{i}}m_1^{t_{i}}m_2^{t_{i}}... m_n^{t_{i}}F[q,q_x,q_{xx},...,q_{nx}]
\ene
To ensure the convergence of this term, the exponential $s$ should be equal or greater than $1$ according to the law of large numbers. When $s=1$, we have
\bee
\begin{aligned}
q(x,\boldsymbol{\bm{\theta}})=&\frac{1}{N} \sum_{k=1}^{N}\boldsymbol{W}^{(1)}_k\sigma_k+\boldsymbol{b}^{(1)}\stackrel{\mathcal{P}}{\rightarrow}\mathbb{E}[\boldsymbol{W}^{(1)}_k\sigma_k]+\boldsymbol{b}^{(1)}=:\mathbb{E}[q],\\
q_{ix}(x,\boldsymbol{\bm{\theta}})=&\frac{1}{N}\sum_{k=1}^{N}\boldsymbol{W}^{(1)}_k\sigma^{(i)}_k\boldsymbol{W}^{(0)i}_k\stackrel{\mathcal{P}}{\rightarrow}\mathbb{E}[\boldsymbol{W}^{(1)}_k\sigma^{(i)}_k\boldsymbol{W}^{(0)i}_k]=:\mathbb{E}[q_i],\quad i=1,...,n\\
\end{aligned}
\ene
Then by the law of large numbers, we have
\bee
\begin{aligned}
\mathbf{K}_{ff}
\stackrel{\mathcal{P}}{\rightarrow}&
F_0\big[\mathbb{E}[q],\mathbb{E}[q_1],\mathbb{E}[q_2],...,\mathbb{E}[q_n]\big]F_0\big[\mathbb{E}[\hat{q}],\mathbb{E}[\hat{q}_1],\mathbb{E}[\hat{q}_2],...,\mathbb{E}[\hat{q}_n]\big].
\end{aligned}
\ene
The first $3N$ terms in above formula have a coefficient $\frac{1}{N^{2s}}$. Therefore, when $s>\frac{1}{2}$, the first $3N$ terms converge to 0.

(b) As every monomial in $\{F_i[q,q_x,q_{xx},...,q_{nx}](x,\bm{\theta})\}_{i=0}^n$ is homogeneous, and \bee
F_0[q,q_x,q_{xx},...,q_{nx}]F_0[\hat{q},\hat{q}_x,\hat{q}_{xx},...,\hat{q}_{nx}]=0,
  \ene
then the convergence is determined by the maximum order term of $F_i$ (The order is defined by the sum of homogeneous parameters $\sum\limits_{l=0}^nt_{il}$ (\ref{homo}).). We assume the number of maximum order is $n_i$ for $F_i$, and we rewrite the $F_i$ as
\bee\label{maxord}
F_i=F^0_i+F^1_i,
\ene
where $F^0_i$ represent the all max order terms in $F_i$, and $F^1_i$ is the remainder terms of $F_i$.
Then, for any two data points $x,\hat{x}$, we have
\bee
\begin{aligned}
\mathbf{K}_{ff}
=&\sum\limits_{k=0}^{3N+1}\bigg(\sum\limits_{i=0}^n\left(F_i^0+F_i^1\right)\frac{\partial q_{ix}(x,\bm{\theta})}{\partial\boldsymbol{\theta}_{k}}\bigg)\bigg(\sum\limits_{j=0}^n\left(\Bar{F}_j^0+\Bar{F}_j^1\right)\frac{\partial q_{jx}(\hat{x},\bm{\theta})}{\partial\boldsymbol{\theta}_{k}}\bigg)\\
=&\sum\limits_{k=0}^{3N+1}\bigg(\sum\limits_{i,j=0}^n\left(F_i^0\Bar{F}_j^0+F_i^1\Bar{F}_j^0+F_i^0\Bar{F}_j^1+F_i^1\Bar{F}_j^1\right)\frac{\partial q_{ix}(x,\bm{\theta})}{\partial\boldsymbol{\theta}_{k}}\frac{\partial q_{jx}(\hat{x},\bm{\theta})}{\partial\boldsymbol{\theta}_{k}}\bigg)\\
\end{aligned}
\ene
When
\bee\label{s1}
s=\frac{2\sum\limits_{l=0}^nt_{i'l}+1}{2\sum\limits_{l=0}^nt_{i'l}+2}=:s_1
\ene
where $\sum\limits_{l=0}^nt_{i'l}=\max\limits_{i=0,...,n}\left\{\sum\limits_{l=0}^nt_{il}\right\}$. By the law of big numbers, we have
\bee
\begin{aligned}
\mathbf{K}_{ff}
=&\bigg(N^{1-2s}\sum\limits_{i,j=0}^n\left(F_i^0\Bar{F}_j^0+F_i^1\Bar{F}_j^0+F_i^0\Bar{F}_j^1+F_i^1\Bar{F}_j^1\right)N^{2s-1}\sum\limits_{k=0}^{3N}\frac{\partial q_{ix}(x,\bm{\theta})}{\partial\boldsymbol{\theta}_{k}}\frac{\partial q_{jx}(\hat{x},\bm{\theta})}{\partial\boldsymbol{\theta}_{k}}\bigg)\\
\stackrel{\mathcal{P}}{\rightarrow}
&F_{i'}^0\Bar{F}_{j'}^0\left(\mathbb{E}\left[\frac{\partial q_{ix}(x,\bm{\theta})}{\partial\boldsymbol{W}^{(0)}_{k}}\frac{\partial q_{jx}(\hat{x},\bm{\theta})}{\partial\boldsymbol{W}^{(0)}_{k}}\right]+
\mathbb{E}\left[\frac{\partial q_{ix}(x,\bm{\theta})}{\partial\boldsymbol{W}^{(1)}_{k}}\frac{\partial q_{jx}(\hat{x},\bm{\theta})}{\partial\boldsymbol{W}^{(1)}_{k}}\right]
+\mathbb{E}\left[\frac{\partial q_{ix}(x,\bm{\theta})}{\partial\boldsymbol{b}^{(0)}_{k}}\frac{\partial q_{jx}(\hat{x},\bm{\theta})}{\partial\boldsymbol{b}^{(0)}_{k}}\right]\right)
\end{aligned}
\ene

(2) $\mathbf{K}_{bb}$: we consider $\mathbf{K}_{bb}$ when $s=\frac{1}{2}$:
\bee
\begin{aligned}
\mathbf{K}_{bb}=&\Big\langle\frac{\partial q_{mx}(x,\bm{\theta})}{\partial\bm{\theta}}^T,\frac{\partial q_{mx}(\hat{x},\bm{\theta})}{\partial\bm{\theta}}\Big\rangle\\
=&\Big\langle\frac{\partial q(x,\bm{\theta})}{\partial\boldsymbol{W}^{(0)}}^T,\frac{\partial q(\hat{x},\bm{\theta})}{\partial\boldsymbol{W}^{(0)}}\Big\rangle
+\Big\langle\frac{\partial q(x,\bm{\theta})}{\partial\boldsymbol{W}^{(1)}}^T,\frac{\partial q(\hat{x},\bm{\theta})}{\partial\boldsymbol{W}^{(1)}}\Big\rangle\\
&+\Big\langle\frac{\partial q(x,\bm{\theta})}{\partial\boldsymbol{b}^{(0)}}^T,\frac{\partial q(\hat{x},\bm{\theta})}{\partial\boldsymbol{b}^{(0)}}\Big\rangle
+\Big\langle\frac{\partial q(x,\bm{\theta})}{\partial\boldsymbol{b}^{(1)}}^T,\frac{\partial q(\hat{x},\bm{\theta})}{\partial\boldsymbol{b}^{(1)}}\Big\rangle
\end{aligned}
\ene

(a) when $m=0$. Recall that
\bee
\bm{q}(x,\bm{\theta})=\frac{1}{N^s}\bm{W}^{(1)}\cdot\sigma(\bm{W}^{(0)}\cdot x+\bm{b}^{(0)})+\bm{b}^{(1)}
\ene
for any two points $x$ and $\hat{x}$
\bee
\begin{aligned}
\mathbf{K}_{bb}=&
\frac{1}{N}\sum_{k=1}^{N}\bigg(\boldsymbol{W}^{(1)2}_k\sigma'_k\hat{\sigma}'_kx\hat{x}\bigg)+
\frac{1}{N}\sum_{k=1}^{N}\sigma_k\hat{\sigma}_k
+\frac{1}{N}\sum_{k=1}^{N}\bigg(\boldsymbol{W}^{(1)2}_k\sigma'_k\hat{\sigma}'_k\bigg)+1\\
\stackrel{\mathcal{P}}{\rightarrow}&
\mathbb{E}\Big[\boldsymbol{W}^{(1)2}_k\sigma'_k\hat{\sigma}'_k\Big]x\hat{x}+\mathbb{E}\Big[\sigma_k\hat{\sigma}_k\Big]+\mathbb{E}\Big[\boldsymbol{W}^{(1)2}_k\sigma_k\hat{\sigma}_k\Big]+1
\end{aligned}
\ene

(b) when $m>0$. Recall that
\bee
\begin{aligned}
q_{mx}(x, \boldsymbol{\bm{\theta}})=&\frac{1}{N^s} \sum_{k=1}^{N}\boldsymbol{W}^{(1)}_k\sigma^{(m)}_k\boldsymbol{W}^{(0)m}_k,
\end{aligned}
\ene

\bee
\begin{aligned}
\mathbf{K}_{bb}=&
\frac{1}{N^{2s}}\sum_{k=1}^{N}\bigg(\boldsymbol{W}^{(1)2}_k\sigma^{(m+1)}_k\hat{\sigma}^{(m+1)}_k\boldsymbol{W}^{(0)2m}_kx\hat{x}
+m^2\boldsymbol{W}^{(1)2}_k\sigma^{(m)}_k\hat{\sigma}^{(m)}_k\boldsymbol{W}^{(0)2(m-1)}_k\\
&+m\boldsymbol{W}^{(1)2}_k\sigma^{(m+1)}_k\hat{\sigma}^{(m)}_k\boldsymbol{W}^{(0)(2m-1)}_kx
+m\boldsymbol{W}^{(1)2}_k\sigma^{(m)}_k\hat{\sigma}^{(m+1)}_k\boldsymbol{W}^{(0)(2m-1)}_k\hat{x}\bigg)\\
&+\frac{1}{N^{2s}}\sum_{k=1}^{N}\bigg(\sigma^{(m)}_k\hat{\sigma}^{(m)}_k\boldsymbol{W}^{(0)2m}_k\bigg)
+\frac{1}{N^{2s}}\sum_{k=1}^{N}\bigg(\boldsymbol{W}^{(1)2}_k\sigma^{(m+1)}_k\hat{\sigma}^{(m+1)}_k\boldsymbol{W}^{(0)2m}_k\bigg)+0\\
\stackrel{\mathcal{P}}{\rightarrow}&
\mathbb{E}\Big[\boldsymbol{W}^{(1)2}_k\sigma^{(m+1)}_k\hat{\sigma}^{(m+1)}_k\boldsymbol{W}^{(0)2m}_k\Big]x\hat{x}
+\mathbb{E}\Big[m^2\boldsymbol{W}^{(1)2}_k\sigma^{(m)}_k\hat{\sigma}^{(m)}_k\boldsymbol{W}^{(0)2(m-1)}_k\Big]\\
&+\mathbb{E}\Big[m\boldsymbol{W}^{(1)2}_k\sigma^{(m+1)}_k\hat{\sigma}^{(m)}_k\boldsymbol{W}^{(0)(2m-1)}_k\Big]x+
\mathbb{E}\Big[m\boldsymbol{W}^{(1)2}_k\sigma^{(m)}_k\hat{\sigma}^{(m+1)}_k\boldsymbol{W}^{(0)(2m-1)}_k\Big]\hat{x}\\
&+\mathbb{E}\Big[\sigma^{(m)}_k\hat{\sigma}^{(m)}_k\boldsymbol{W}^{(0)2m}_k\Big]+
\mathbb{E}\Big[\boldsymbol{W}^{(1)2}_k\sigma^{(m+1)}_k\hat{\sigma}^{(m+1)}_k\boldsymbol{W}^{(0)2m}_k\Big]
\end{aligned}
\ene

(3) $\mathbf{K}_{fb}(=\mathbf{K}_{bf}^T)$:
discussion similar to (1):

(a) When  $F_0[q,q_x,q_{xx},...q_{nx}]\neq0$ or exist a monomial in the polynomial in $\{F_i[q,q_x,q_{xx},...q_{nx}](x,\bm{\theta})\}_{i=1}^n$ is non-homogeneous.
To ensure the convergence of this term, the exponential s should be equal or greater than 1 according to the law of large numbers. Therefore, when $s=1$, for any two data points $x$, $\hat{x}$, we have
\bee
\begin{aligned}
\mathbf{K}_{fb}=
&\Big\langle\nabla_{\bm{\theta}}\mathcal{F}[q](x,\bm{\theta}), \nabla_{\bm{\theta}}q_{mx}(\hat{x},\bm{\theta})\Big\rangle\\
=&\sum\limits_{k=0}^{3N}\bigg(\sum\limits_{i=0}^nF_i[q,q_x,q_{xx},...,q_{nx}]\frac{\partial q_{ix}(x,\bm{\theta})}{\partial\boldsymbol{\theta}_{k}}\bigg)\bigg(\frac{\partial q_{mx}(x,\bm{\theta})}{\partial\boldsymbol{\theta}_{k}}\bigg)\\
&+F_0[q,q_x,q_{xx},...q_{nx}]\frac{\partial q_{mx}(x,\bm{\theta})}{\partial\boldsymbol{b}^{(1)}}\\
\stackrel{\mathcal{P}}{\rightarrow}&
\begin{cases}
F_0\big[\mathbb{E}[q],\mathbb{E}[q_1],\mathbb{E}[q_2],...,\mathbb{E}[q_n]\big],\quad m=0,\\
0,\qquad\qquad\qquad\qquad\qquad\qquad\quad m>0.
\end{cases}
\end{aligned}
\ene

(b) When $F_0[q,q_x,q_{xx},...q_{nx}]=0$ and every monomial in the polynomials in $\{F_i[q,q_x,q_{xx},...,q_{nx}](x,\bm{\theta})\}_{i=1}^n$ is homogeneous. To balance the coefficient of the neural network, we set
\bee\label{s2}
s=\frac{\sum\limits_{l=0}^nt_{i'l}+1}{\sum\limits_{l=0}^nt_{i'l}+2}=:s_2
\ene
($\sum\limits_{l=0}^nt_{i'l}=\max\limits_{i=0,...,n}\left\{\sum\limits_{l=0}^nt_{il}\right\}$ and $\{t_{il}\}$ are defined in (\ref{homo})). And for any $i,j\in\{0,1,...,n\}$, we define
\bee
\begin{aligned}
N^{2s-1}\bigg(\frac{\partial q_{ix}(x,\bm{\theta})}{\partial\boldsymbol{\theta}}\bigg)^T\bigg(\frac{\partial q_{jx}(x,\bm{\theta})}{\partial\boldsymbol{\theta}}\bigg)
=& \frac{1}{N}\sum\limits_{k=0}^{3N+1}N^s\bigg(\frac{\partial q_{ix}(x,\bm{\theta})}{\partial\boldsymbol{\theta}_k}\bigg)^T N^s\bigg(\frac{\partial q_{jx}(x,\bm{\theta})}{\partial\boldsymbol{\theta}_k}\bigg)\\
\stackrel{\mathcal{P}}{\rightarrow}&\mathbb{E}[q_i,q_j]
\end{aligned}
\ene
Then, we have
\bee
\begin{aligned}
\mathbf{K}_{fb}=
&\sum\limits_{k=0}^{3N}\bigg(\sum\limits_{i=0}^nF_i[q,q_x,q_{xx},...,q_{nx}]\frac{\partial q_{ix}(x,\bm{\theta})}{\partial\boldsymbol{\theta}_{k}}\bigg)\bigg(\frac{\partial q_{mx}(x,\bm{\theta})}{\partial\boldsymbol{\theta}_{k}}\bigg)\\
=&\sum\limits_{i=0}^nN^{1-2s}(F_i^0+F_i^1)N^{2s-1}\sum\limits_{k=0}^{3N}\frac{\partial q_{ix}(x,\bm{\theta})}{\partial\boldsymbol{\theta}_{k}}\frac{\partial q_{mx}(x,\bm{\theta})}{\partial\boldsymbol{\theta}_{k}}\\
\stackrel{\mathcal{P}}{\rightarrow}&
F_{i'}^0\mathbb{E}[q_i,q_j]
\end{aligned}
\ene

In summary, $\mathbf{K}_{bb}$ convergence by probability when
$s\geq\frac{1}{2}$. If
\bee F_0[q,q_x,q_{xx},...,q_{nx}]F_0[\hat{q},\hat{q}_x,\hat{q}_{xx},...,\hat{q}_{nx}]\neq0
\ene ($F_0[q,q_x,q_{xx},...,q_{nx}]\neq0$) or exist a monomial in the polynomial in $\{F[q,q_x,q_{xx},...q_{nx}](x,\bm{\theta})\}_{i=1}^n$ is non-homogeneous (case A), $\mathbf{K}_{ff}$ ($\mathbf{K}_{bf}$) convergence
by probability when $s\geq1$. And if $F_0[q,q_x,q_{xx},...,q_{nx}]F_0[\hat{q},\hat{q}_x,\hat{q}_{xx},...,\hat{q}_{nx}]=0$ ($F_0[q,q_x,q_{xx},...,q_{nx}]=0$) or every monomial in the polynomials in $\{F_i[q,q_x,q_{xx},...,q_{nx}](x,\bm{\theta})\}_{i=1}^n$ is homogeneous (case B), $\mathbf{K}_{ff}$ ($\mathbf{K}_{bf}$) convergence
by probability when $s\geq s_1$(\ref{s1})
($s\geq s_2$(\ref{s2})).

In other words, in case A, $\mathbf{K}$ convergence by probability when $s\geq1$. In case B, $\mathbf{K}$ convergence by probability when $s\geq s_1$. $\qedsymbol$

\v \noindent {\bf Appendix C. \, Proof of Theorem 2.3}\label{Th2}

\v Before we prove Theorem 2.3, we first prove a couple of lemmas. We add a discussion of the general PDE based on the original proof.

{\bf Lemma C.1} under the condition of Theorem 2.3, we have
\bee
\begin{aligned}
\sup\limits_{t\in [0,T]}\left\|\frac{\partial q_{mx}}{\partial \boldsymbol{\theta}}\right\|_{\infty}=&\mathcal{O}(\frac{1}{N^s})\\
\sup\limits_{t\in [0,T]}\left\|\frac{\partial \mathcal{F}[q]}{\partial \boldsymbol{\theta}}\right\|_{\infty}=&\mathcal{O}(\frac{1}{N^s})
\end{aligned}
\ene
where $\bm{\theta}=\{\bm{W}^{(0),T},\bm{b}^{(0),T},\bm{W}^{(1)},\bm{b}^{(1)}\}$.

{\it Proof.} We still assume $\mathcal{F}[q](x,\bm{\theta})=F[q,q_x,q_{xx},...q_{nx}](x,\bm{\theta})=f(x)$. recall that
\bee
\begin{aligned}
q(x, \boldsymbol{\bm{\theta}})=&\frac{1}{N^s} \sum_{k=1}^{N}\boldsymbol{W}^{(1)}_k\sigma\left(\boldsymbol{W}^{(0)}_kx
+\boldsymbol{b}^{(0)}_k\right)+\boldsymbol{b}^{(1)},\\
q_{mx}(x, \boldsymbol{\bm{\theta}})=&\frac{1}{N^s} \sum_{k=1}^{N}\boldsymbol{W}^{(1)}_k\sigma^{(m)}\boldsymbol{W}^{(0)m}_k,
\end{aligned}
\ene
and
\bee
\begin{aligned}
\frac{\partial \mathcal{F}[q](x,\bm{\theta})}{\partial\bm{\theta}_{k}}
=&\sum\limits_{i=0}^nF_i[q,q_x,q_{xx},...q_{nx}]\frac{\partial q_{ix}(x,\bm{\theta})}{\partial\bm{\theta}_{k}}.
\end{aligned}
\ene
By the Uniformly boundedness of weight, derivative of activation function $\sigma^{(k)}$ and $F_i$ (assumptions (i), (ii), (iv)), We have
\bee
\begin{aligned}
\sup\limits_{t\in [0,T]}\left\|\frac{\partial \mathcal{F}[q]}{\partial \bm{\theta}_{k}}\right\|_{\infty}
=&\sup\limits_{t\in [0,T]}\left\|\sum\limits_{i=0}^nF_i[q,q_x,q_{xx},...q_{nx}]\frac{\partial q_{ix}(x,\bm{\theta})}{\partial\bm{\theta}_{k}}\right\|_{\infty}\\
\leq&\sum\limits_{i=0}^n\sup\limits_{t\in [0,T]}\left\|F_i[q,q_x,q_{xx},...q_{nx}]\right\|_{\infty}\sup\limits_{t\in [0,T]}\left\|\frac{\partial q_{ix}(x,\bm{\theta})}{\partial\bm{\theta}_{k}}\right\|_{\infty}\\
\leq&nC\frac{C}{N^s}=O(\frac{1}{N^s}), \quad k=1,2,...,3N+1.
\end{aligned}
\ene
This completes the proof.
$\qedsymbol$

{\bf Lemma C.2} under the condition of Theorem 2.3, we have
\bee
\begin{aligned}
\lim\limits_{N\rightarrow\infty}\sup\limits_{t\in[0,T]}\left\|\frac{1}{N^s}(\bm{\theta}(t)-\bm{\theta}(0))\right\|_2=0
\end{aligned}
\ene
{\it Proof.}
\bee
\mathcal{L}(\bm{\theta})=\mathcal{L}_f(\bm{\theta})+\mathcal{L}_b(\bm{\theta})=\displaystyle\frac{1}{2}\sum_{j=1}^{N_f}\left|\mathcal{F}[q](x^j_f,\bm{\theta})-f(x^j_f)\right|^2+\displaystyle\frac{1}{2}\sum_{j=1}^{N_b}\left|q_{mx}(x_b^j,\bm{\theta})-g(x^j_b)\right|^2,
\ene
Because of
\bee
\begin{aligned}
\frac{d\bm{\theta}}{dt}=&-\nabla_{\bm{\theta}}\mathcal{L}(\bm{\theta}),
\end{aligned}
\ene
we have
\bee
\begin{aligned}
&\left\|\frac{1}{N^s}(\bm{\theta}(t)-\bm{\theta}(0))\right\|_{2}
=\left\|\frac{1}{N^s} \int_{0}^{t} \frac{d \bm{\theta}(\tau)}{d \tau} d \tau\right\|_{2}
=\left\|\frac{1}{N^s} \int_{0}^{t} \frac{\partial \mathcal{L}(\bm{\theta}(\tau))}{\partial \bm{\theta}} d \tau\right\|_{2}\\
&=\left\|\frac{1}{N^s}\int_{0}^{t}\bigg[\sum_{j=1}^{N_f}(\mathcal{F}[q](x^j_f,\bm{\theta})-f(x^j_f))\frac{\partial\mathcal{F}[q](x^j_f,\bm{\theta})}{\partial \bm{\theta}}+\sum_{j=1}^{N_b}(\frac{\partial^mq(x_b^j,\bm{\theta})}{\partial x^m}-g(x^j_b))\frac{\partial q_{mx}(x_b^j,\bm{\theta})}{\partial \bm{\theta}}\bigg]d \tau\right\|_{2}\\
&\leq \boldsymbol{A}_1+\boldsymbol{A}_2
\end{aligned}
\ene
where
\bee
\begin{aligned}
\boldsymbol{A}_1
&=\left\|\frac{1}{N^s}\int_{0}^{t}\bigg[\sum_{j=1}^{N_f}(\mathcal{F}[q](x^j_f,\bm{\theta})-f(x^j_f))\frac{\partial\mathcal{F}[q](x^j_f,\bm{\theta})}{\partial \bm{\theta}}\bigg]d \tau\right\|_{2}\\
&\leq\frac{1}{N^s}\int_{0}^{t}\left\| \alpha\sum_{j=1}^{N_f}(\mathcal{F}[q](x^j_f,\bm{\theta})-f(x^j_f))\frac{\partial\mathcal{F}[q](x^j_f,\bm{\theta})}{\partial \bm{\theta}}\right\|_{2}d \tau\\
&=\frac{1}{N^s}\int_{0}^{t}\sqrt{\sum_{k=1}^{3N+1}\left( \sum_{j=1}^{N_f}(\mathcal{F}[q](x^j_f,\bm{\theta})-f(x^j_f))\frac{\partial\mathcal{F}[q](x^j_f,\bm{\theta})}{\partial \boldsymbol{W}^{(l)}_{k}}\right)^2}d \tau\\
&\leq\frac{1}{N^s}\int_{0}^{t}\left\|\frac{\partial\mathcal{F}[q](x^j_f,\bm{\theta})}{\partial \boldsymbol{W}^{(l)}_{k}}\right\|_{\infty}\sqrt{\sum_{k=1}^{3N+1}\left( \sum_{j=1}^{N_f}(\mathcal{F}[q](x^j_f,\bm{\theta})-f(x^j_f))\right)^2}d \tau\\
&=\frac{C}{N^{s-1/2}}\int_{0}^{t}\left\|\frac{\partial\mathcal{F}[q](x^j_f,\bm{\theta})}{\partial \boldsymbol{W}^{(l)}_{k}}\right\|_{\infty} \left|\sum_{j=1}^{N_f}\left(\mathcal{F}[q](x^j_f,\bm{\theta})-f(x^j_f)\right)\right|d \tau\\
&=\mathcal{O}(\frac{1}{N^{2s-1/2}}).
\end{aligned}
\ene

In the same way, we have
\bee
\begin{aligned}
\boldsymbol{A}_2
&=\left\|\frac{1}{N^s}\int_{0}^{t}\bigg[ \sum_{j=1}^{N_b}(\frac{\partial^mq(x_b^j,\bm{\theta})}{\partial x^m}-g(x^j_b))\frac{\partial q_{mx}(x_b^j,\bm{\theta})}{\partial \boldsymbol{W}^{(l)}}\bigg]d \tau\right\|_{2}=\mathcal{O}(\frac{1}{N^{2s-1/2}}).
\end{aligned}
\ene
$\qedsymbol$

{\bf Lemma C.3} Under the condition of Theorem 2.3, when $s>\frac{1}{4}$ we have
\bee
\begin{aligned}
\lim\limits_{N\rightarrow\infty}\sup\limits_{t\in[0,T]}\left\|\frac{1}{N^s}(\sigma^{(k)}|_{t=T}-\sigma^{(k)}|_{t=0})\right\|_2=0
\end{aligned}
\ene
{\it Proof.} by the mean-value theorem for $\sigma^{(k)}$ and Lemma C.2 .
$\qedsymbol$

\v {\bf Lemma C.4} under the condition of Theorem 2.3, we have
\bee
\begin{aligned}
\lim\limits_{N\rightarrow\infty}\sup\limits_{t\in[0,T]}\left\|\frac{\partial q_{mx}(x,\bm{\theta}(t))}{\partial \bm{\theta}}-\frac{\partial q_{mx}(x,\bm{\theta}(0))}{\partial \bm{\theta}}\right\|_2=0\\
\lim\limits_{N\rightarrow\infty}\sup\limits_{t\in[0,T]}\left\|\frac{\partial \mathcal{F}[q](x,\bm{\theta}(t))}{\partial \bm{\theta}}-\frac{\partial \mathcal{F}[q](x,\bm{\theta}(0))}{\partial \bm{\theta}}\right\|_2=0
\end{aligned}
\ene
{\it Proof.} Recall that
\bee
\begin{aligned}
\frac{\partial \mathcal{F}[q](x,\bm{\theta})}{\partial\bm{\theta}_{k}}
=&\sum\limits_{i=0}^nF_i[q,q_x,q_{xx},...q_{nx}]\frac{\partial q_{ix}(x,\bm{\theta})}{\partial\bm{\theta}_{k}}.
\end{aligned}
\ene
Take $\bm{W}^{(0)}_k$ for example, we have
\bee
\begin{aligned}
&\sup\limits_{t\in[0,T]}\left\|\frac{\partial \mathcal{F}[q](x,\bm{\theta}(t))}{\partial\boldsymbol{W}^{(0)}_{k}}-\frac{\partial \mathcal{F}[q](x,\bm{\theta}(t))}{\partial\boldsymbol{W}^{(0)}_{k}}\right\|_2\\
&=\sup\limits_{t\in[0,T]}\left\|\sum\limits_{i=0}^nF_i[q,q_x,q_{xx},...q_{nx}]\frac{\partial q_{ix}(x,\bm{\theta}(t))}{\partial\boldsymbol{W}^{(0)}_{k}}-\sum\limits_{i=0}^nF_i[q,q_x,q_{xx},...q_{nx}]\frac{\partial q_{ix}(x,\bm{\theta}(0))}{\partial\boldsymbol{W}^{(0)}_{k}}\right\|_2\\
&=\sup\limits_{t\in[0,T]}\Bigg\|\sum\limits_{i=0}^nF_i[q,q_x,q_{xx},...,q_{nx}]\bigg(\Big(\frac{1}{N^s}\boldsymbol{W}^{(1)}_k(t)\sigma^{(i+1)}_k(t)\boldsymbol{W}^{(0)i}_k(t)x+\frac{i}{N^s}\boldsymbol{W}^{(1)}_k(t)\sigma^{(i)}_k(t)\boldsymbol{W}^{(0)(i-1)}_k(t)\Big) \\
&\qquad -\Big(\frac{1}{N^s}\boldsymbol{W}^{(1)}_k(0)\sigma^{(i+1)}_k(0)\boldsymbol{W}^{(0)i}_k(0)x+\frac{i}{N^s}\boldsymbol{W}^{(1)}_k(0)\sigma^{(i)}_k(0)\boldsymbol{W}^{(0)(i-1)}_k(0)\Big)\bigg)\Bigg\|_2 \\
&\leq\sup\limits_{t\in[0,T]}\left\|\sum\limits_{i=0}^nF_i[q,q_x,q_{xx},...,q_{nx}]\Big(\frac{1}{N^s}\boldsymbol{W}^{(1)}_k(t)\sigma^{(i+1)}_k(t)\boldsymbol{W}^{(0)i}_k(t)x-\frac{1}{N^s}\boldsymbol{W}^{(1)}_k(0)\sigma^{(i+1)}_k(0)\boldsymbol{W}^{(0)i}_k(0)x\Big)\right\|_2\\
&\qquad +\sup\limits_{t\in[0,T]}\left\|\sum\limits_{i=0}^nF_i[q,q_x,q_{xx},...,q_{nx}]\Big(\frac{i}{N^s}\boldsymbol{W}^{(1)}_k(t)\sigma^{(i)}_k(t)\boldsymbol{W}^{(0)(i-1)}_k(t)-\frac{i}{N^s}\boldsymbol{W}^{(1)}_k(0)\sigma^{(i)}_k(0)\boldsymbol{W}^{(0)(i-1)}_k(0)\Big)\right\|_2\\
&=A_1+A_2.
\end{aligned}
\ene
Then

\bee\label{A1}
\begin{aligned}
A_1&=\sup\limits_{t\in[0,T]}\left\|\sum\limits_{i=0}^nF_i[q,q_x,q_{xx},...,q_{nx}]\Big(\frac{1}{N^s}\boldsymbol{W}^{(1)}_k(t)\sigma^{(i+1)}_k(t)\boldsymbol{W}^{(0)i}_k(t)x-\frac{1}{N^s}\boldsymbol{W}^{(1)}_k(0)\sigma^{(i+1)}_k(0)\boldsymbol{W}^{(0)i}_k(0)x\Big)\right\|_2\\
&\leq\sup\limits_{t\in[0,T]}\bigg\|\sum\limits_{i=0}^nF_i[q,q_x,q_{xx},...,q_{nx}]\frac{1}{N^s}\boldsymbol{W}^{(1)}_k(t)\sigma^{(i+1)}_k(t)\Big(\boldsymbol{W}^{(0)i}_k(t)-\boldsymbol{W}^{(0)i}_k(0)\Big)x\bigg\|_2\\
&\qquad +\sup\limits_{t\in[0,T]}\bigg\|\sum\limits_{i=0}^nF_i[q,q_x,q_{xx},...,q_{nx}]\frac{1}{N^s}\Big(\boldsymbol{W}^{(1)}_k(t)\sigma^{(i+1)}_k(t)-\boldsymbol{W}^{(1)}_k(0)\sigma^{(i+1)}_k(0)\Big)\boldsymbol{W}^{(0)i}_k(0)x\bigg\|_2\\
&\leq\sum\limits_{i=0}^n\sup\limits_{t\in[0,T]}\bigg\|F_i[q,q_x,q_{xx},...,q_{nx}]\bigg\|_2\bigg\|\boldsymbol{W}^{(1)}_k(t)\bigg\|_2\bigg\|\sigma^{(i+1)}_k(t)\bigg\|_2\bigg\|\frac{1}{N^s}\Big(\boldsymbol{W}^{(0)i}_k(t)-\boldsymbol{W}^{(0)i}_k(0)\Big)x\bigg\|_2\\
&\qquad +\sum\limits_{i=0}^n\sup\limits_{t\in[0,T]}\bigg\|F_i[q,q_x,q_{xx},...,q_{nx}]\bigg\|_2\bigg\|\frac{1}{N^s}\Big(\boldsymbol{W}^{(1)}_k(t)\sigma^{(i+1)}_k(t)-\boldsymbol{W}^{(1)}_k(0)\sigma^{(i+1)}_k(0)\Big)\bigg\|_2\bigg\|\boldsymbol{W}^{(0)i}_k(0)x\Big\|_2\\
\end{aligned}
\ene
Because
\bee
\begin{aligned}
&\bigg\|\frac{1}{N^s}\Big(\boldsymbol{W}^{(1)}_k(t)\sigma^{(i+1)}_k(t)-\boldsymbol{W}^{(1)}_k(0)\sigma^{(i+1)}_k(0)\Big)\bigg\|_2\\
\leq&\bigg\|\frac{1}{N^s}\Big(\boldsymbol{W}^{(1)}_k(t)\sigma^{(i+1)}_k(t)-\boldsymbol{W}^{(1)}_k(t)\sigma^{(i+1)}_k(0)\Big)\bigg\|_2+\bigg\|\frac{1}{N^s}\Big(\boldsymbol{W}^{(1)}_k(t)\sigma^{(i+1)}_k(0)-\boldsymbol{W}^{(1)}_k(0)\sigma^{(i+1)}_k(0)\Big)\bigg\|_2\\
\leq&\frac{1}{N^s}\bigg\|\boldsymbol{W}^{(1)}_k(t)\bigg\|_2\bigg\|\sigma^{(i+1)}_k(t)-\sigma^{(i+1)}_k(0)\bigg\|_2+\frac{1}{N^s}\bigg\|\boldsymbol{W}^{(1)}_k(t)-\boldsymbol{W}^{(1)}_k(0)\bigg\|_2\bigg\|\sigma^{(i+1)}_k(0)\bigg\|_2\\
=&\mathcal{O}(\frac{1}{N^{2s-1/2}}).
\end{aligned}
\ene
Then we have
\bee
A_1=\mathcal{O}(\frac{1}{N^{2s-1/2}}).
\ene
By the same way
\bee
\begin{aligned}
A_2=\mathcal{O}(\frac{1}{N^{2s-1/2}}).
\end{aligned}
\ene
and the same method can be used to $\boldsymbol{W}^{(1)}_{k},\boldsymbol{b}^{(0)}_{k},\boldsymbol{b}^{(1)}_{k}$. Therefore
\bee
\begin{aligned}
\left\|\frac{\partial \mathcal{F}[q](x,\bm{\theta}(t))}{\partial \bm{\theta}}-\frac{\partial \mathcal{F}[q](x,\bm{\theta}(0))}{\partial \bm{\theta}}\right\|_2=\mathcal{O}(\frac{1}{N^{2s-1/2}}).\\
\left\|\frac{\partial q_{mx}(x,\bm{\theta}(t))}{\partial \bm{\theta}}-\frac{\partial q_{mx}(x,\bm{\theta}(0))}{\partial \bm{\theta}}\right\|_2=\mathcal{O}(\frac{1}{N^{2s-1/2}}).
\end{aligned}
\ene
$\qedsymbol$

When the above-mentioned four Lemmas are finished, the raw theorem can be proved.

{\it Proof of Theorem 2.3.} The Kernel matrix $K$ can be divided into the multiply of the Jacobian matrix:
\bee
\mathbf{K}(t)=\begin{bmatrix}
\mathbf{K}_{ff}(t) & \mathbf{K}_{fb}(t) \\
\mathbf{K}_{bf}(t) & \mathbf{K}_{bb}(t)
\end{bmatrix}=\begin{bmatrix}
\mathbf{J}_{f}(t) \\
\mathbf{J}_{b}(t)
\end{bmatrix}\begin{bmatrix}
\mathbf{J}^T_{f}(t), & \mathbf{J}^T_{b}(t)
\end{bmatrix}:=\mathbf{J}(t)\mathbf{J}^T(t),
\ene
where $\mathbf{J}_{f}(t)$
\bee
\begin{aligned}
\mathbf{J}_{f}(t)=\Big(\nabla_{\bm{\theta}}\mathcal{F}[q](x^i_f,\bm{\theta})\Big)_{i=1,...,N_f}\in\mathbb{R}^{N_f\times(3N+1)},\\
\mathbf{J}_{b}(t)=\Big(\nabla_{\bm{\theta}}q_{mx}(x^i_b,\bm{\theta})\Big)_{i=1,...,N_b}\in\mathbb{R}^{N_b\times(3N+1)},\\
\end{aligned}
\ene
Then we have
\bee
\begin{aligned}
\left\|\mathbf{K}(t)-\mathbf{K}(0)\right\|_2
=&\left\|\mathbf{J}(t)\mathbf{J}^T(t)-\mathbf{J}(0)\mathbf{J}^T(0)\right\|_2\\
\leq&\left\|(\mathbf{J}(t)-\mathbf{J}(0))\mathbf{J}^T(t)\right\|_2+\left\|\mathbf{J}(0)(\mathbf{J}^T(t)-\mathbf{J}^T(0))\right\|_2\\
\leq&\left\|\mathbf{J}(t)-\mathbf{J}(0)\right\|_2\left\|\mathbf{J}^T(t)\right\|_2+\left\|\mathbf{J}(0)\right\|_2
\left\|\mathbf{J}^T(t)-\mathbf{J}^T(0)\right\|_2.
\end{aligned}
\ene
By Lemma C.1, $\left\|\mathbf{J}^T(t)\right\|_F$ and $\left\|\mathbf{J}(0)\right\|_F$ are bounded. And because of the equivalence of norm, $\left\|\mathbf{J}^T(t)\right\|_2$ and $\left\|\mathbf{J}(0)\right\|_2$ are bounded.

Then, we consider the convergence of $\left\|\mathbf{J}^T(t)-\mathbf{J}^T(0)\right\|_2$. By Lemma C.4, it is easy to see that
\bee
\begin{aligned}
\left\|\mathbf{J}(t)-\mathbf{J}(0)\right\|_F^2
=&\sum\limits_{i=1}^{N_f}\left\|\frac{\partial \mathcal{F}[q](x_i,\bm{\theta}(t))}{\partial \bm{\theta}}-\frac{\partial \mathcal{F}[q](x_i,\bm{\theta}(0))}{\partial \bm{\theta}}\right\|_F^2\\
&+\sum\limits_{i=1}^{N_b}\left\|\frac{\partial q_{mx}(x,\bm{\theta}(t))}{\partial \bm{\theta}}-\frac{\partial q_{mx}(x,\bm{\theta}(0))}{\partial \bm{\theta}}\right\|_F^2\\
=&\mathcal{O}(\frac{1}{N^{4s-1}}).
\end{aligned}
\ene
Thus, $\left\|\mathbf{J}(t)-\mathbf{J}(0)\right\|_F$ is converge to 0 when $N\rightarrow\infty$. And because of the equivalence of norm, $\left\|\mathbf{J}(t)-\mathbf{J}(0)\right\|_2$ is converge to 0 when $N\rightarrow\infty$.

Then we have $\left\|\mathbf{K}(t)-\mathbf{K}(0)\right\|_2\rightarrow0$, when $N\rightarrow\infty$.
$\qedsymbol$


\v \v \noindent {\bf Acknowledgement}

The work  was supported by the National Natural Science Foundation of China  under Grant Nos. 11925108 and 12226332.

\end{document}